\documentclass[10pt]{article}

\usepackage[a4paper,margin=0.75in]{geometry}
\usepackage{microtype}
\usepackage{times}

\usepackage{amsmath,amssymb,amsfonts}
\usepackage{algorithm}
\usepackage{algorithmic}
\usepackage{siunitx}
\usepackage{booktabs}
\usepackage{multirow}
\usepackage{tabularx,array}

\usepackage{graphicx}
\usepackage{caption}
\usepackage{subcaption}
\usepackage{xcolor}

\usepackage{authblk}
\usepackage{abstract}

\usepackage[numbers,sort&compress]{natbib}
\setcitestyle{numbers,square}

\usepackage{hyperref}
\hypersetup{
  colorlinks=true,
  linkcolor=blue,
  citecolor=blue,
  urlcolor=blue
}

\title{Robust and Explainable Bicuspid Aortic Valve Diagnosis Using Stacked Ensembles on Echocardiography}

\author[1,*]{Christos Chrysanthos Nikolaidis}
\author[2]{Vasileios Sachpekidis}
\author[3]{Nikolas Moustakidis}
\author[4]{Theofilos Moustakidis}
\author[1]{Pavlos S. Efraimidis}

\affil[1]{Department of Electrical and Computer Engineering, Democritus University of Thrace, Xanthi, 67100, Greece}
\affil[2]{Department of Cardiology, Papageorgiou Hospital, Thessaloniki, Greece}
\affil[3]{Department of Informatics, Aristotle University of Thessaloniki, Thessaloniki, Greece}
\affil[4]{Department of Bioinformatics, University of Thessaly, Larissa, Greece}
\affil[*]{Corresponding author: \texttt{cnikolai@ee.duth.gr}}

\date{}

\begin{document}

\twocolumn[
\begin{@twocolumnfalse}
\maketitle

\begin{abstract}
Transthoracic echocardiography (TTE) is the first-line imaging modality for diagnosing bicuspid aortic valve (BAV), yet diagnostic performance varies with operator expertise and image quality. We developed an explainable AI model that distinguishes BAV from tricuspid aortic valves (TAV) using routinely acquired parasternal long-axis (PLAX) cine loops. A multi-backbone video ensemble was trained and evaluated using a leakage-aware, stratified outer cross-validation protocol on $N{=}90$ patient studies (48 BAV, 42 TAV). Across fixed outer splits and 10 random seeds, the calibrated stacked ensemble achieved an outer-CV F1-score of $0.907$ and recall of $0.877$. Frame-level Grad-CAM localized salient evidence to the aortic root and leaflet plane, while globally aggregated SHAP values quantified each video backbone's contribution to the stacked prediction, enabling transparent, case-level auditability. These findings indicate that PLAX-based video ensembles can support reliable BAV/TAV classification from routine echocardiographic cine loops and may facilitate earlier detection in non-specialist or resource-limited clinical settings.
\end{abstract}

\vspace{0.5em}
\vspace{1em}
\end{@twocolumnfalse}
]

\section{Introduction}
\label{sec:intro}

Bicuspid aortic valve (BAV) is the most common congenital valvular malformation, affecting 1--2\% of the general population and associated with elevated risks of aortic stenosis, regurgitation, infective endocarditis, and aortopathy~\citep{BravermanCheng2021BAV}. Although some individuals remain asymptomatic for many years, a large proportion eventually develop clinically significant valve dysfunction or aortic complications, which can lead to heart failure, sudden cardiac death, or the need for early surgery. Early and reliable identification of BAV is therefore crucial, as timely surveillance and intervention can substantially reduce morbidity and mortality over the lifelong course of the disease.

Transthoracic echocardiography (TTE) is the primary diagnostic modality for BAV diagnosis, but its accuracy depends heavily on operator expertise and image acquisition quality. Routine TTE often exhibits variability when compared with magnetic resonance imaging (MRI) or surgical findings, and non-expert readers frequently misclassify valve morphology~\citep{hillebrand2017accuracy, ayad2011echoaccuracy}. Although the parasternal short-axis (PSAX) is the preferred view for determining cusp number and establishing the diagnosis of BAV, it is often difficult to acquire, particularly for inexperienced users.  In contrast, the parasternal long-axis (PLAX) view is easier to obtain even in non-specialist or resource-limited settings---and may still provide diagnostically useful cues (e.g.  leaflet doming, hinge-line thickening, and aortic root geometry) that raise suspicion of BAV presence. This makes PLAX an attractive target for point-of-care and community-based screening, where high-quality PSAX imaging may not be consistently feasible. However, these cues are often subtle, highlighting the need for automated, interpretable decision-support tools that can generalize beyond expert operators.

Despite rapid advances in deep learning for cardiac image analysis, several gaps limit the deployment of reliable BAV/TAV classification models for routine PLAX cine loops. \textbf{(i)} Many prior works rely on single architectures or still-frame pipelines that fail to capture the spatio–temporal signatures of valve motion~\citep{chen2020echocnn}. \textbf{(ii)} Although video-based architectures show promise in related echocardiography tasks~\citep{ouyang2020video, hong2022hypertrophic} and broader ultrasound classification where motion cues are essential~\citep{howard2020improving}, to our knowledge, no prior PLAX-based BAV studies have combined multiple pretrained video backbones into a rigorously evaluated ensemble. \textbf{(iii)} Ensemble stacking on small datasets is vulnerable to target leakage at the meta-layer unless out-of-fold (OOF) construction is used; at the same time, clinically deployable systems require transparent, case-level explanations that connect spatial evidence with model reasoning~\citep{selvaraju2017gradcam, lundberg2017shap, vandervelden2022xaiReview}. Related PLAX-based detection of valvular disease, such as severe aortic stenosis, demonstrates the feasibility of video-based echocardiographic modeling~\citep{holste2023severeAS}. In medical imaging more broadly, ensemble approaches improve robustness when designed and evaluated carefully~\citep{muller2022ensemble, ganaie2022ensembleReview, ganie2025heartdisease}, and recent self-supervised or transfer-learning strategies further enhance performance on small datasets~\citep{zhou2024echosmallDL}.

In this study, we introduce a reproducible multi-backbone video ensemble for BAV/TAV classification from PLAX cine loops that leverages inner-CV out-of-fold (OOF) base probabilities to train diverse meta-learners under fixed, strictly held-out outer folds in a leakage-aware, stratified outer cross-validation protocol. On a curated dataset of \( N{=}90 \) DICOM studies (48 BAV / 42 TAV), the final stacked and calibrated ensemble achieves an outer-CV Accuracy of \(0.894\), F1-score of \(0.907\), Precision of \(0.941\), Recall of \(0.877\), AUROC of \(0.849\), AP of \(0.885\), and Brier score of \(0.149\). The evaluation protocol is summarized in Section~\ref{sec:method}, and full metrics and calibration analyses are reported in Section~\ref{sec:results}.

Our contributions are summarized as follows:
\begin{itemize}
    \item \textbf{We present a leakage-aware video ensemble effective on limited data.} We demonstrate reliable BAV/TAV classification from routine PLAX cine loops in a small, single-center cohort using a leakage-aware, stratified outer cross-validation design, supporting high-sensitivity screening applications.
    \item \textbf{PLAX multi-backbone video ensemble.} We introduce a multi-backbone stacked ensemble pipeline (MC3, R3D, X3D, R2P1D, S3D) fused via compact meta-learners trained on 10-dimensional base probabilities derived from inner-CV out-of-fold predictions.
    \item \textbf{Leakage-aware stacking and calibration.} We implement meta-training using inner-CV OOF base predictions under fixed outer splits, ensuring the outer-test fold remains strictly unseen until final evaluation, and apply post-hoc calibration to improve probability reliability.
    \item \textbf{Dual-level explainability and auditability.} We combine frame-level Grad-CAM, which localizes salient spatial evidence to the aortic root and leaflet plane, with globally aggregated SHAP explanations that quantify the contribution of each video backbone to the stacked prediction, alongside per-sample logging to enable transparent, case-wise auditing and risk communication.
\end{itemize}

The remainder of this paper is organized as follows. Section~\ref{sec:related} reviews related work on PLAX-based analysis, video modeling, stacking ensembles, and explainability. Section~\ref{sec:data} describes the dataset and preprocessing, including temporal normalization and region-of-interest (ROI) handling. Section~\ref{sec:method} presents the proposed leakage-aware stacking framework, base backbones, meta-learners, calibration, and explainability tools. Section~\ref{sec:results} reports overall performance, ROC/PR and calibration analyses, explainability findings, and error analysis. Section~\ref{sec:discussion} discusses implications, limitations, and future directions, and Section~\ref{sec:conclusion} concludes the paper.

\section{Related Work}
\label{sec:related}

The detection of bicuspid aortic valve (BAV) on transthoracic echocardiography (TTE) remains challenging due to operator dependence and variable view quality. Comparative studies with surgical findings or magnetic resonance imaging (MRI) highlight substantial variability in routine TTE and frequent misclassification by non-experts~\citep{hillebrand2017accuracy, ayad2011echoaccuracy}. These limitations motivate automated systems that exploit diagnostically informative cues present in easily acquired parasternal long-axis (PLAX) sequences.

Deep learning has advanced cardiac image analysis across a wide range of tasks, from chamber segmentation to disease detection. On PLAX cine, automated detection of severe aortic stenosis has achieved strong performance~\citep{holste2023severeAS}. For valve morphology and related tasks on \emph{static} frames, CNN- and U-Net--based architectures have reached high accuracy~\citep{chen2020echocnn}. However, frame-based models often neglect spatio--temporal dynamics critical for valve motion analysis, limiting their ability to generalize across acquisition variability or subtle leaflet morphologies.

To capture motion, several studies have used 3D CNNs and transformer-based encoders for echocardiographic sequences~\citep{ouyang2020video,hong2022hypertrophic}. Similar temporal modeling strategies have also improved ultrasound video classification, where motion cues are essential~\citep{howard2020improving}. Hybrid CNN--transformer ensembles have recently shown improved generalization for medical video classification by integrating local texture and global temporal context~\citep{lee2023hybridEnsemble}. Nevertheless, PLAX-based BAV/TAV classification using \emph{multi-backbone} video ensembles, rigorously evaluated under leakage-aware protocols, remains underexplored.

Ensemble learning improves robustness and generalization, with consistent gains reported for stacking, bagging, and boosting across modalities and domains~\citep{ganaie2022ensembleReview, muller2022ensemble}. Cardiovascular imaging applications also benefit from carefully designed ensembles and meta-learners~\citep{ganie2025heartdisease}. For small datasets, preventing target leakage at the meta-level is critical: meta-learners should be trained on out-of-fold (OOF) base predictions so that meta-features remain strictly out-of-sample~\citep{ganaie2022ensembleReview}. These practices motivate our leakage-aware OOF stacking with fixed outer splits, ensuring that evaluation metrics reflect true generalization.

Explainability remains essential for clinical adoption. Grad-CAM provides intuitive saliency maps for convolutional networks~\citep{selvaraju2017gradcam}, while SHAP offers model-agnostic feature attribution~\citep{lundberg2017shap}. Recent reviews emphasize combining localized visual explanations with quantitative attribution to support clinical auditing and trust in medical imaging~\citep{vandervelden2022xaiReview}. However, applying both frame-level (Grad-CAM) and meta-level (SHAP) explainability to video-ensemble pipelines for valve morphology remains largely unexplored.

Although the present study is centralized and single-institutional, privacy-preserving multi-center learning remains an important direction for future clinical deployment. Federated learning enables collaborative model development across institutions without centralizing raw patient data, which is particularly relevant for medical imaging applications where data sharing is constrained by privacy and governance requirements. In this work, we focus on developing and evaluating a leakage-aware, calibrated, and explainable PLAX-based classification pipeline; its extension to distributed clinical settings is discussed in Section~\ref{sec:future_work}.

In summary, previous studies have advanced static and video-based cardiac image analysis, ensemble learning, and model interpretability. However, PLAX-based BAV versus TAV classification using rigorously leakage-aware, explainable multi-backbone video ensembles remains unaddressed. The proposed approach addresses this gap by integrating motion-aware backbones, OOF stacking, and dual-level explanations to enhance the reliability and transparency of valve morphology classification.

\section{Dataset and Preprocessing}
\label{sec:data}

\subsection{Dataset and labeling}
\label{sec:dataset}
The dataset comprised \(N = 90\) anonymized transthoracic echocardiographic studies acquired in the parasternal long-axis (PLAX) view during routine clinical care at a single, high volume valvular heart disease clinic (Papageorgiou General Hospital, Thessaloniki, Greece). Patients provided informed consent for use of anonymized data. Studies were exported in DICOM format and manually labeled by an expert cardiologist with significant experience in valvular heart disease, as either bicuspid (BAV; \(n = 48\)) or tricuspid aortic valve (TAV; \(n = 42\)); see Table~\ref{tab:dataset}. Labels were assigned by the interpreting cardiologist based on detailed review of all available echocardiographic views, incorporating valve morphology, cusp fusion pattern, prior imaging (e.g., transoesophageal echocardiography), and longitudinal follow-up, in accordance with standardized criteria~\cite{Michelena2021BAVConsensusEJCTS}. Studies with indeterminate morphology, inadequate visualization of the aortic valve, or severe artifacts were excluded. Each patient contributed a single PLAX cine loop to avoid within-patient correlation. All cine loops were acquired during routine clinical scanning without specialized imaging protocols. Variation in frame rate, depth, sector width, and gain was retained to reflect real-world practice. 

\subsection{Preprocessing pipeline}
\label{sec:preproc}

\paragraph{DICOM decoding and frame extraction.}
\label{sec:dicom}
PLAX cine loops were loaded from DICOM format, and individual frames were extracted together with the available metadata. Frame-level temporal ordering was retained exactly as acquired to respect variations in frame rate and study duration. After DICOM decoding, frames were represented as 8-bit grayscale raster images (intensity range \(0\)--\(255\)). No spatial transformations or intensity normalization were applied at this stage.

\begin{figure}[!tbp]
  \centering
  \includegraphics[width=\columnwidth]{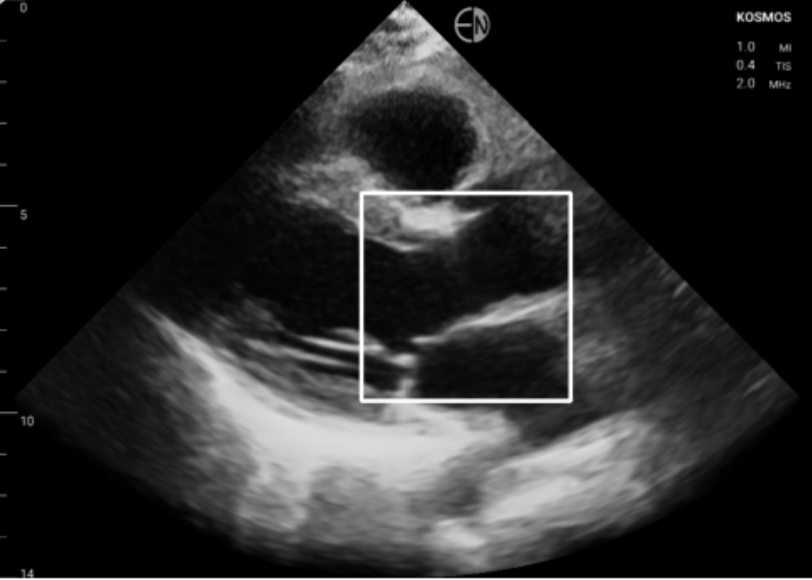}
  \caption{Representative PLAX frame with the fixed ROI \((x,y,w,h)=(420,275,240,240)\).
  After cropping, frames are resized to \(224\times224\); zero padding is used only if the ROI extends beyond image bounds. The procedure is per-frame and label-agnostic.}
  \label{fig:roi}
\end{figure}

\paragraph{ROI extraction and resizing.}
\label{sec:roi}
A coarse label-agnostic region of interest (ROI) was defined to include the aortic valve, left ventricular outflow tract, and the proximal aortic root. As illustrated in Fig.~\ref{fig:roi}, we used a fixed square ROI in raster coordinates
\[
r = (x,y,w,h) = (420,\,275,\,240,\,240),
\]
chosen empirically to provide consistent coverage of the aortic root among patients. The ROI coordinates were selected once on a small label-agnostic development subset prior to cross-validation and then fixed for all folds and seeds. For each frame, the corresponding crop \(I_r\in\mathbb{R}^{h\times w}\) was extracted and resized to \(224\times224\) pixels using bilinear interpolation to match the input requirements of the pretrained video backbones (MC3, R3D, X3D, R2P1D, S3D). Because the ROI is square, resizing does not distort aspect ratio; if the ROI extended beyond image bounds (rare), we used zero padding with a constant value of 0.

\begin{table}[!tbp]
  \centering\small
  \caption{Summary of the PLAX transthoracic echocardiography cohort for BAV/TAV classification.}
  \label{tab:dataset}
  \begin{tabular}{lcc}
    \toprule
    Class & \# Studies & Proportion \\
    \midrule
    TAV & 42 & 46.7\% \\
    \addlinespace[0.35em]
    BAV & 48 & 53.3\% \\
    \midrule
    Total & 90 & 100\% \\
    \bottomrule
  \end{tabular}
\end{table}

\paragraph{Temporal normalization.}
\label{sec:temporal}
Cine loops varied substantially in length between patients (typically 80--120 frames). To construct fixed-length video tensors, each PLAX sequence was normalized to a target length of \(T^\ast=85\) frames. Let a raw clip contain \(T\) frames. We form a \(T^\ast\)-frame clip by truncation when \(T\ge T^\ast\) and by repeating the last available frame when \(T<T^\ast\):
\[
\mathbf{F} =
\begin{cases}
(f_1,\ldots,f_{85}), & T \ge 85,\\[4pt]
(f_1,\ldots,f_T,\;\underbrace{f_T,\ldots,f_T}_{85-T}), & T<85.
\end{cases}
\]
This deterministic procedure avoids sampling randomness in a small dataset while ensuring uniform tensor shapes across the backbones. We selected $T^\ast = 85$ as a compromise between retaining diagnostically relevant valve motion and enforcing a uniform sequence length for all backbones. Notably, no study in our cohort required last-frame repetition padding (0/90 cases), as the first 85 frames consistently captured one to two complete cardiac cycles. This deterministic windowing preserves the aortic valve opening--closing dynamics without requiring ECG synchronization.

\paragraph{Intensity mapping and channel handling.}
\label{sec:intensity}
All PLAX frames are native single-channel grayscale images. Pixel intensities were first converted to floating point and linearly scaled to the \([0,1]\) range by division by 255. To match the input format required by pretrained video backbones (which expect three RGB channels), each grayscale frame was tiled into three identical channels. No dataset-dependent or ImageNet/Kinetics mean--std normalization was applied; models were trained directly on the three-channel \([0,1]\) inputs. After preprocessing, each study is represented as a tensor of shape
\[
(C,T^\ast,H^\ast,W^\ast) = (3,\,85,\,224,\,224).
\]
\paragraph{Training-time augmentation.}
\label{sec:augment}
To improve generalization in a limited-data setting, light spatial augmentations were applied on-the-fly during training. These included small in-plane rotations (\(\pm 5^{\circ}\)) and slight spatial translations. Augmentations were intentionally conservative to avoid generating anatomically implausible configurations or altering valve morphology cues. No temporal augmentations were used, preserving consistent cardiac motion dynamics across minibatches. We did not use horizontal or vertical flipping for augmentation. PLAX views are typically acquired in a standardized orientation, so flipping can create anatomically implausible laterality and weaken learned spatial priors. In a preliminary ablation, horizontal flipping did not improve performance and slightly worsened probability calibration.

\begin{table}[!tbp]
\centering
\small
\caption{Preprocessing configuration used in all experiments.}
\label{tab:preproc_config}
\setlength{\tabcolsep}{4pt}
\renewcommand{\arraystretch}{1.1}
\begin{tabularx}{\columnwidth}{@{}lX@{}}
\toprule
\textbf{Component} & \textbf{Setting} \\
\midrule
View & PLAX cine loop (one study per patient) \\
\addlinespace[0.35em]
ROI crop (x,y,w,h) & (420, 275, 240, 240) \\
\addlinespace[0.35em]
Resize & 224$\times$224 (bilinear) \\
\addlinespace[0.35em]
Target clip length $T^\ast$ & 85 frames \\
\addlinespace[0.35em]
Temporal padding & Repeat last frame (when $T < T^\ast$) \\
\addlinespace[0.35em]
Temporal truncation & Keep first $T^\ast$ frames (when $T \ge T^\ast$) \\
\addlinespace[0.35em]
Intensity scaling & divide by 255 (8-bit, 0--255) $\rightarrow [0,1]$ \\
\addlinespace[0.35em]
Channels & grayscale $\rightarrow$ 3-channel repeat \\
\addlinespace[0.35em]
Mean--std normalization & none \\
\addlinespace[0.35em]
Augmentation (train only) & small rotations ($\pm5^\circ$) + translations \\
\bottomrule
\end{tabularx}
\end{table}

\subsection{Leakage-safe preprocessing for out-of-fold stacking}
\label{sec:oof_preproc}
All preprocessing operations were deterministic and applied identically to training and test samples within each outer fold, without using any statistics from the held-out outer-test split. Inner cross-validation then produced out-of-fold (OOF) base predictions, and these OOF predictions formed the inputs for meta-learner training, ensuring that all meta-features were strictly out-of-sample. All preprocessing hyperparameters in Table~\ref{tab:preproc_config} were fixed \emph{a priori} and kept identical across folds and seeds; no outer-test information was used to adjust ROI placement, clip length, or normalization.

\subsection{Logging for explainability and error analysis}
\label{sec:logging}
For each study and each outer fold, preprocessing metadata, model predictions, and backbone outputs were logged. This enabled case-wise explainability using (i) frame-level Grad-CAM heatmaps computed on the input clip, and (ii) global SHAP attribution on meta-features derived from stacked base probabilities. Misclassified cases were saved for subsequent qualitative error analysis (Section~\ref{sec:results}).

\section{Methodology}\label{sec:method}
\subsection{Overview}
\begin{figure*}[!ht]
  \centering
  \includegraphics[width=\linewidth]{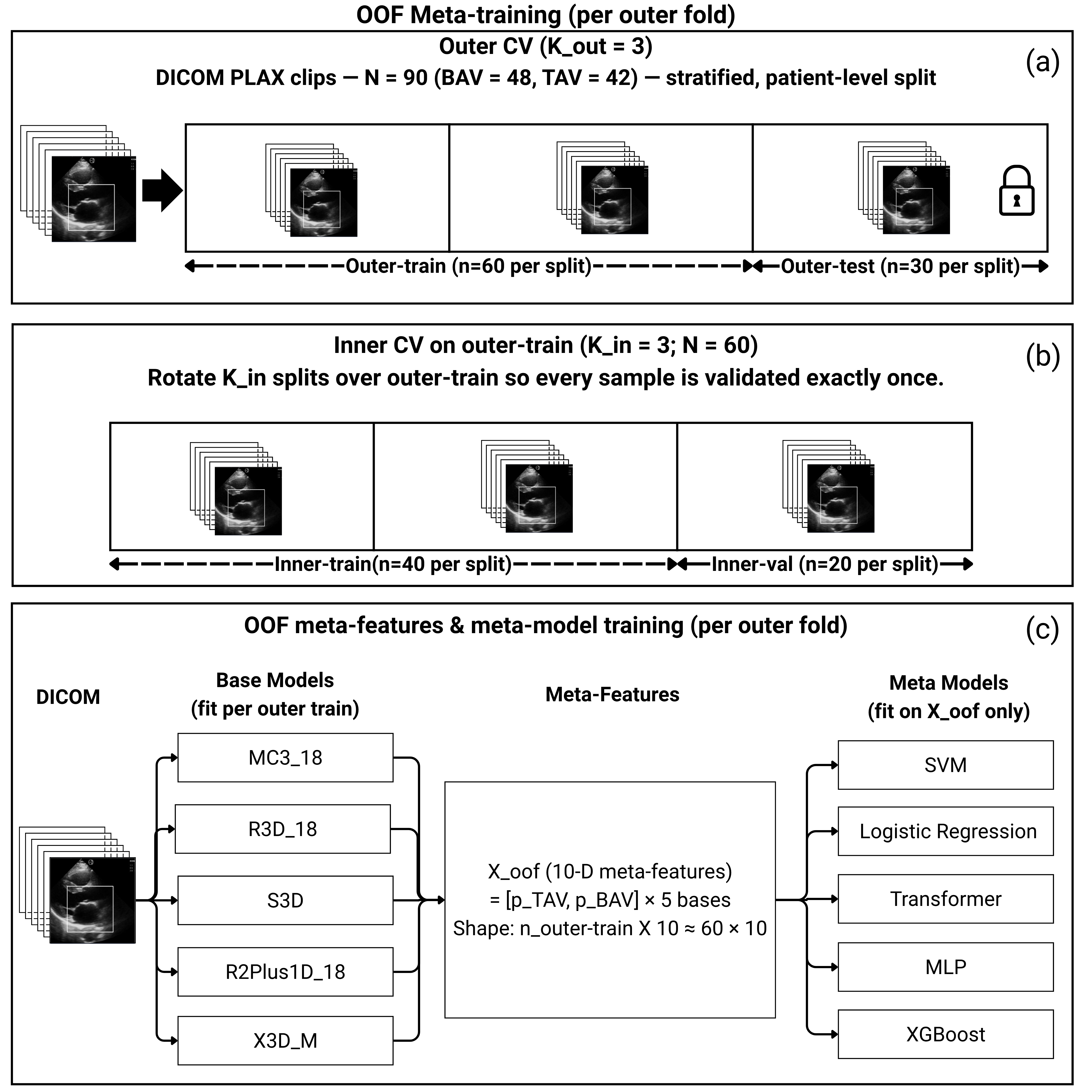}
  \caption{\textbf{OOF meta-training per outer fold.} \textbf{(a) Outer cross-validation (} $K_{\text{out}}{=}3$; patient-level, stratified\textbf{)}: each fold holds out $n_{\text{outer-test}}\!\approx\!30$ studies, leaving $n_{\text{outer-train}}\!=\!60$ for training. \textbf{(b) Inner CV on the outer-train (} $K_{\text{in}}{=}3$\textbf{)} rotates three inner folds so every sample is validated exactly once. \textbf{(c) OOF meta-features and meta-model training:} base models trained on each inner-train predict softmax probabilities on inner-val; concatenating $[p_{\text{TAV}},p_{\text{BAV}}]$ across five bases forms 10-D meta-features $X_{\text{oof}}$. Meta-learners see only OOF features; the outer-test split is never used for any fitting or selection.}
  \label{fig:oof_meta_training}
\end{figure*}

\begin{figure*}[!ht]
  \centering
  \includegraphics[width=\linewidth]{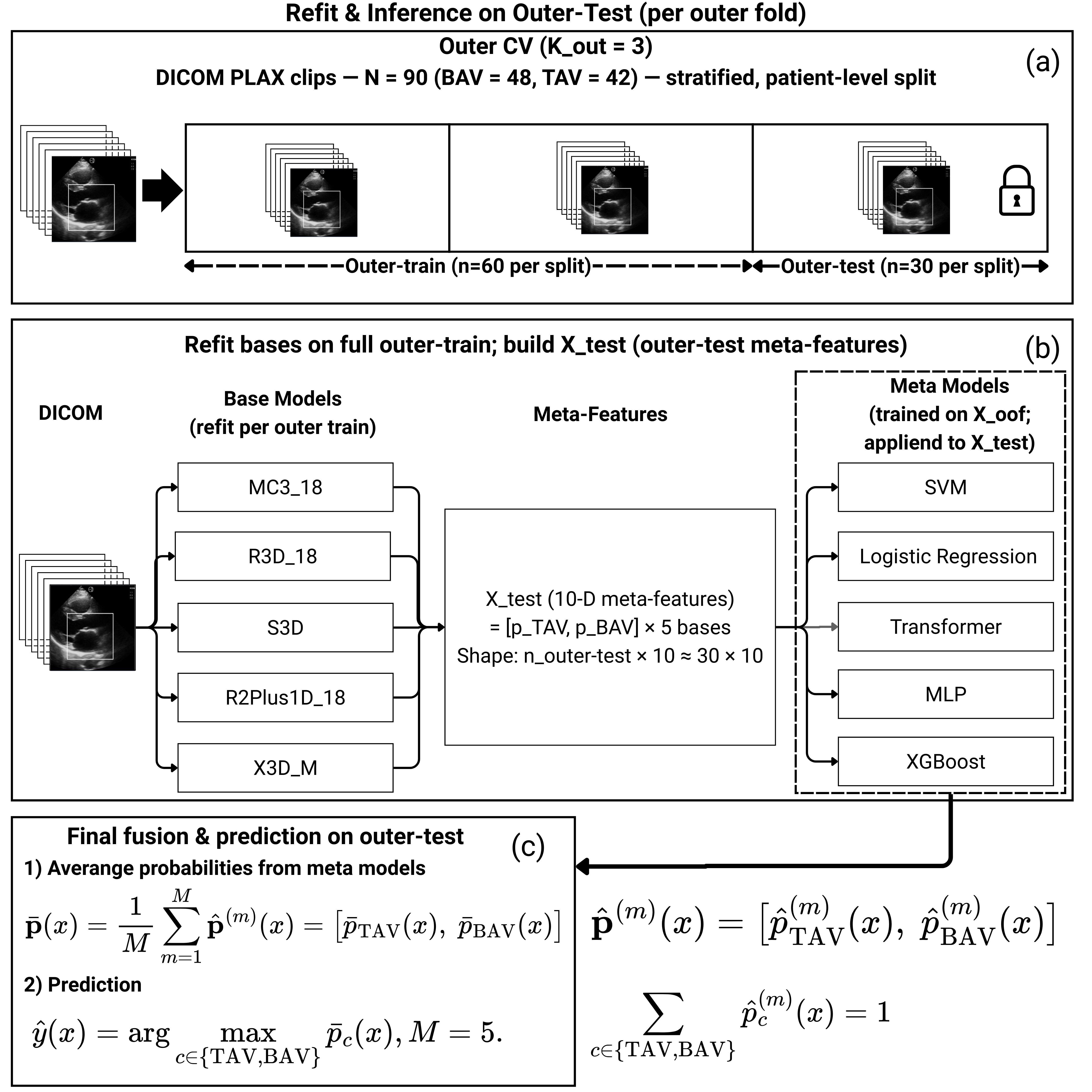}
  \caption{\textbf{Refit \& inference per outer fold.} (a) Outer CV at the patient level ($K_{\text{out}}{=}3$). (b) Refit each base on the full outer-train; score the outer-test to build $X_{\text{test}}$ (10-D meta-features per sample), then apply meta-models trained on $X_{\text{oof}}$. (c) Average meta probabilities to obtain the final prediction. Meta parameters are those fitted on $X_{\text{oof}}$; no refitting or tuning uses the outer-test split.}
  \label{fig:inference_flow}
\end{figure*}

We adopt a two-level architecture trained under a fully nested cross-validation framework. At the \emph{base layer}, five pretrained video backbones independently process each PLAX cine loop and output class probabilities for TAV and BAV. These ten probabilities are concatenated into a 10-dimensional meta-feature vector, which serves as input to a diverse set of meta-learners forming the \emph{meta layer}. The ordering of backbones and the within-backbone probability order $[p_{\text{TAV}}, p_{\text{BAV}}]$ are fixed to ensure consistent feature semantics across folds and seeds. The OOF-generation and meta-training workflow is illustrated in Fig.~\ref{fig:oof_meta_training}, while the refit-and-inference stage is shown in Fig.~\ref{fig:inference_flow}. A step-by-step description of the nested stacking protocol is provided in Algorithm~\ref{alg:clean_stacking}. During evaluation, base models are refit on each outer-train split, their probabilities on the corresponding outer-test split are stacked, and the meta-learners output an ensemble BAV probability that is subsequently calibrated (Section~\ref{sec:calibration}) and thresholded for classification.

Stacking these heterogeneous base and meta models is intended to hedge against model misspecification at both levels, allowing the ensemble to exploit complementary spatio--temporal representations and decision rules while remaining compatible with the limited sample size.

\subsection{Base video models and training}\label{sec:base_models}
Five pretrained video backbones were used: MC3, R3D, X3D, R2P1D, and S3D, initialized from Kinetics-400 weights (\(K=400\)). Each input is a \((C,T,H,W)=(3,85,224,224)\) tensor derived from grayscale PLAX frames by ROI cropping, resizing, tiling to three channels, and \([0,1]\) intensity scaling (Section~\ref{sec:data}); no mean--std normalization was applied. All backbones were fine-tuned with Adam (learning rate \(10^{-4}\); batch size \(4\); 70 epochs maximum) using cross-entropy loss. BAV was treated as the positive class for probability reporting and threshold-based metric computation. Early stopping based on the inner-validation macro-averaged F1-score (macro-F1; patience = 20) mitigated overfitting in the inner loops. Given the near-balanced class distribution (48/42), no class weighting was used. Data augmentation was applied only during training within the outer-train split (including inner-CV training runs) and never to outer-test samples (Section~\ref{sec:preproc}).

We retained the original Kinetics classification head and appended a lightweight task adapter. Let the pretrained head produce pre-softmax logits \(z\in\mathbb{R}^{K}\) over the \(K\) pretraining classes (\(K=400\)). The adapter is a linear layer with parameters \(W\in\mathbb{R}^{2\times K}\) and \(b\in\mathbb{R}^{2}\) that maps \(z\) to task logits \(u=Wz+b\), followed by a softmax to obtain \(p(y\mid x)=\mathrm{softmax}(u)\) for TAV vs.\ BAV. The backbone parameters, the original pretrained head, and the adapter parameters were jointly optimized during fine-tuning; the adapter parameters \((W,b)\) were randomly initialized. In preliminary inner-CV ablations (within each outer-training partition), the logit-adapter design yielded slightly better overall validation performance (across standard metrics) than replacing the pretrained classifier with a randomly initialized 2-class head on penultimate features; therefore, we used it uniformly across all backbones and folds.

These five video backbones were selected to span complementary 3D-CNN design families while remaining computationally tractable on a small medical dataset. MC3 and R3D provide mixed and fully 3D convolutional baselines, respectively, with different trade-offs between spatial and temporal receptive fields. R2P1D uses (2+1)D factorization, separating spatial and temporal kernels to improve optimization, whereas S3D employs separable convolutions that reduce parameter count while preserving spatio--temporal capacity. X3D is a lightweight, width/depth-expanded architecture optimized for video efficiency. We exploit Kinetics pretraining to initialize generic motion and texture filters before fine-tuning on PLAX echocardiography.

\subsection{Leakage-aware inference pipeline}
\label{sec:clean_stacking}
We use a fully nested, leakage-aware evaluation protocol. Stratified patient-level outer cross-validation (\(K_{\text{out}} = 3\)) yields three held-out test folds. Within each outer-train split, stratified inner CV (\(K_{\text{in}}=3\)) produces out-of-fold (OOF) base predictions for every outer-train instance. Concretely, in each inner fold the five video backbones are trained on the inner-train subset and used to predict two-class softmax probabilities on the corresponding inner-validation subset. Let \(p^{(b)}(x)\in[0,1]^2\) denote the class-probability vector (TAV, BAV) from backbone \(b\). For sample \(x\), the stacked meta-feature is
\[
\phi(x)= [p^{(1)}(x); \dots; p^{(5)}(x)] \in \mathbb{R}^{10}.
\]
Concatenating predictions across inner folds forms the OOF matrix
\[
X_{\text{oof}}\in\mathbb{R}^{N_{\text{train}}\times 10},\qquad y_{\text{train}}\in\{0,1\}^{N_{\text{train}}},
\]
where each row is generated by base models that did not train on that sample. After meta-training, each backbone is refit on the full outer-train split and applied to the outer-test split to produce \(X_{\text{test}}\in\mathbb{R}^{N_{\text{test}}\times 10}\). Meta-models trained on \(X_{\text{oof}}\) then score \(X_{\text{test}}\); their predicted probabilities are averaged to form the final ensemble probability. The outer-test labels are used only for the final evaluation. The complete protocol is given in Algorithm~\ref{alg:clean_stacking} and summarized in the accompanying figures.

All experiments use stratified patient-level outer cross-validation with $K_{\text{out}}=3$ folds. Within each
outer-train split, stratified inner cross-validation with $K_{\text{in}}=3$ folds is used to generate out-of-fold (OOF) base predictions for stacking. Meta-level calibration uses an additional stratified cross-fitting step with $K_{\text{cal}}=3$ folds on $(X_{\text{oof}}, y_{\text{train}})$ to produce meta-OOF ensemble scores for Platt scaling (Section~\ref{sec:calibration}). The performance reported summarizes 10 independent random seeds under fixed outer splits, where seeds affect optimization and initialization but not the composition of the outer folds.

\begin{algorithm}[t]
\caption{OOF stacking with fixed outer splits}
\label{alg:clean_stacking}
\begin{algorithmic}[1]
\REQUIRE Stratified outer CV ($K_{\text{out}}{=}3$); inner CV ($K_{\text{in}}{=}3$); base models $\mathcal{B}$; meta-models $\mathcal{M}$; labels $y$
\FOR{each outer fold with outer-train indices $\mathcal{I}_{\text{train}}$ and outer-test indices $\mathcal{I}_{\text{test}}$}
  \STATE Set $y_{\text{train}} \leftarrow y[\mathcal{I}_{\text{train}}]$.
  \STATE Initialize $X_{\text{oof}}\in\mathbb{R}^{|\mathcal{I}_{\text{train}}|\times 10}$ to zeros.
  \STATE Define an index map $\pi:\mathcal{I}_{\text{train}}\rightarrow\{1,\ldots,|\mathcal{I}_{\text{train}}|\}$.
  \FOR{each inner split of $\mathcal{I}_{\text{train}}$ into $(\mathcal{I}_{\text{inner-train}},\mathcal{I}_{\text{inner-val}})$}
    \FOR{each base $b\in\mathcal{B}$}
      \STATE Train $b$ on inner-train data indexed by $\mathcal{I}_{\text{inner-train}}$.
      \STATE Predict softmax probabilities on $\mathcal{I}_{\text{inner-val}}$ and write them to the two columns of $X_{\text{oof}}$ assigned to base $b$.
    \ENDFOR
  \ENDFOR
  \STATE Train each meta-model $m\in\mathcal{M}$ on $(X_{\text{oof}}, y_{\text{train}})$.
  \STATE Refit each base $b\in\mathcal{B}$ on the full outer-train split $\mathcal{I}_{\text{train}}$.
  \STATE Use the refit bases to compute $X_{\text{test}}\in\mathbb{R}^{|\mathcal{I}_{\text{test}}|\times 10}$ on the outer-test split.
  \STATE Obtain \emph{meta-OOF} ensemble scores on $X_{\text{oof}}$ via stratified cross-fitting of the meta-models (e.g., $K_{\text{cal}}{=}3$), averaging the five meta-model probabilities within each held-out calibration split, so each training score is produced by meta-models that did not train on that instance.
  \STATE Fit a recalibration function (Platt scaling)~\citep{platt1999probabilistic} on these meta-OOF scores versus $y_{\text{train}}$.
  \STATE Apply the fitted Platt recalibrator to the meta-OOF scores and select a decision threshold $\tau^\star$ on the outer-train split $\mathcal{I}_{\text{train}}$ (maximize F1 over $\{0.000,0.001,\ldots,1.000\}$; if tied, select the smallest $\tau$).
  \STATE Score $X_{\text{test}}$ using meta-models trained on the full $X_{\text{oof}}$, apply the fitted recalibrator, and threshold with $\tau^\star$ to obtain final predictions on the outer-test split.
\ENDFOR
\end{algorithmic}
\end{algorithm}

\subsection{Meta-learners and hyperparameter search}\label{sec:meta_models}
Five complementary meta-learner families are trained on the 10-D OOF feature vectors: logistic regression (LR), support vector machines (SVM), a shallow multilayer perceptron (MLP), gradient-boosted trees (XGBoost), and a compact Transformer encoder operating on the 10-D token sequence. Each meta-feature vector consists of the concatenated base-model softmax outputs (TAV/BAV probabilities) in a fixed order, yielding a consistent semantic mapping across folds and random seeds.

For LR, MLP, and SVM, we standardize features with \texttt{Standard\allowbreak Scaler} \emph{within} cross-validation, fitting the scaler only on the meta-training portion of each split and applying it to the corresponding validation data. This prevents leakage of fold statistics (mean and variance) into validation. Tree-based (XGBoost) and Transformer meta-models use the raw 10-D probability features without scaling.

For each outer fold, hyperparameters for LR, MLP, XGBoost, and SVM were selected via randomized 3-fold cross-validation on $X_{\mathrm{oof}}$, using the macro-averaged F1-score (macro-F1) as the model-selection criterion, since in binary classification a single F1-score is often interpreted with respect to one designated positive class, whereas macro-F1 averages the class-specific F1-scores for BAV and TAV and therefore promotes balanced performance across both valve morphologies. For SVM, the kernel type (RBF or polynomial) was included in the search space, yielding one selected SVM configuration per outer fold. After hyperparameter selection, the chosen configuration for each meta-learner, together with the fitted scaler when applicable, was refit on the full $(X_{\mathrm{oof}}, y_{\mathrm{train}})$ of the corresponding outer fold and then applied to $X_{\mathrm{test}}$. Importantly, the outer-test split was not used for hyperparameter selection, preprocessing estimation, threshold selection, or calibration.

We further employed a Transformer-based meta-classifier within the stacking stage. Rather than operating on video frames, this model takes as input the concatenated class-probability outputs of the five fine-tuned base video models, yielding 10 meta-features per study (TAV/BAV probabilities from each backbone). Each scalar meta-feature is treated as a token and linearly projected into a $d$-dimensional embedding space. A learnable classification token (\texttt{[CLS]}) is prepended to the token sequence, and the resulting sequence is processed by a multi-layer Transformer encoder with learnable positional embeddings. The final BAV probability is produced from the \texttt{[CLS]} representation through a small feed-forward prediction head. This design enables the meta-classifier to capture non-linear interactions and dependencies among base-model outputs, improving the aggregation of complementary or conflicting evidence across backbones. To reduce overfitting risk in this low-dimensional regime, the Transformer uses a small fixed architecture chosen \emph{a priori} and is trained for 70 epochs (fixed across outer folds and seeds), without fold-specific architectural tuning.

For each outer fold, the final ensemble probability is obtained by averaging the predicted BAV probabilities of the five selected meta-models (LR, SVM, MLP, XGBoost, Transformer) evaluated on $X_{\text{test}}$.

The chosen meta-learners intentionally cover diverse inductive biases in a low-dimensional (10-D) feature regime. Logistic regression provides a strongly regularized linear baseline with well-understood calibration properties. Kernel SVMs (kernel type selected between RBF and polynomial) capture smooth non-linear decision boundaries with tight control of model capacity. The shallow MLP adds a flexible but compact neural non-linearity, while XGBoost models higher-order feature interactions and is robust to moderate feature scaling differences. Finally, the compact Transformer treats the 10 base probabilities as a short token sequence and can learn subtle cross-backbone dependencies that are not easily captured by pointwise models. Averaging across these heterogeneous meta-learners reduces reliance on any single inductive bias and improves robustness of the final ensemble.

\subsection{Probability calibration}
\label{sec:calibration}
Probability calibration is applied \emph{only} at the meta level and is designed to avoid in-sample bias at the stacking layer. Within each outer fold, after constructing $X_{\text{oof}}$ and selecting meta-model hyperparameters, we produce \emph{meta-out-of-fold} (meta-OOF) ensemble scores on the outer-train set via an additional stratified cross-fitting step with $K_{\text{cal}}{=}3$ folds on $(X_{\text{oof}}, y_{\text{train}})$. Concretely, for each calibration split, the meta-models are trained on the meta-train partition and used to predict probabilities on the held-out meta-validation partition; averaging the five meta-model probabilities yields one \emph{out-of-sample} ensemble score per training instance. Collecting these scores across the $K_{\text{cal}}$ splits yields a meta-OOF score vector $\tilde{p}\in[0,1]^{N_{\text{train}}}$ in which every element is produced by meta-models that did not train on that instance. We then fit a logistic recalibration function (Platt scaling)~\citep{platt1999probabilistic} mapping $\tilde{p}$ to calibrated probabilities by minimizing log-loss on $(\tilde{p}, y_{\text{train}})$.

The calibrated meta-OOF training probabilities are subsequently used to select the decision threshold $\tau^\star$ without accessing outer-test labels (Section~\ref{sec:threshold}).

At inference time, meta-models trained on the full $X_{\text{oof}}$ score the outer-test features $X_{\text{test}}$, and the fitted Platt recalibrator is applied to these scores. Calibration is applied to the final averaged meta-ensemble probability (not to individual base backbones). Outer-test labels are never used for fitting base models, meta-models, recalibration, or threshold selection. Calibration quality is reported using the Brier score~\citep{brier1950verification}, reliability diagrams (5 bins), and expected calibration error (ECE; 5 bins) on the held-out outer-test folds.

\subsection{Threshold selection for classification}\label{sec:threshold}
All reported point metrics (Accuracy, Precision, Recall, F1) require a decision threshold $\tau$ on the calibrated BAV probability. To avoid any outer-test leakage, $\tau$ is selected \emph{within each outer fold} using only outer-train information. We used macro-F1 for meta-model selection to promote balanced performance across BAV and TAV, whereas the operating threshold $\tau^\star$ was chosen to maximize the conventional positive-class F1 because the intended use is screening-oriented and treats BAV as the positive class.

Concretely, after fitting the Platt recalibrator on the \emph{meta-OOF} ensemble scores (Section~\ref{sec:calibration}),
we apply that recalibrator to the meta-OOF scores to obtain calibrated training probabilities
$\{\hat{p}_i\}_{i\in\mathcal{I}_{\text{train}}}$. We then choose the threshold
\[
\tau^\star = \arg\max_{\tau\in\mathcal{T}} \mathrm{F1}\big(\mathbb{I}[\hat{p}_i \ge \tau],\, y_i\big),
\]
where $\mathcal{T}$ is a fixed grid (e.g., $\{0.000, 0.001, \ldots, 1.000\}$). If multiple thresholds achieve the same maximum F1 on the grid, we select the smallest threshold (which typically favors higher Recall) to reflect the screening-oriented objective. The selected $\tau^\star$ is then applied \emph{once} to the calibrated outer-test probabilities to produce final binary predictions. At no point are outer-test labels used to set $\tau$.

In small cohorts, the selected threshold $\tau^\star$ may vary across folds due to finite-sample variability and fold-specific calibration shifts. For transparency, we report $\tau^\star$ per fold (mean~$\pm$~SD across 10 seeds) in Table~\ref{tab:fold_thresholds}.

\subsection{Explainability}\label{sec:explainability}
We provide explainability at two complementary levels: (i) \emph{base-level} saliency maps from Grad-CAM for each backbone, and (ii) \emph{meta-level} SHAP attributions over the stacked 10-D meta-features. All analyses on the outer-test split use models trained without accessing that fold.

\paragraph{Base-level Grad-CAM}\label{sec:gradcam}
For each backbone we compute class-discriminative saliency maps using Grad-CAM~\citep{selvaraju2017gradcam}, applied to the deepest 3D convolutional block that preserves spatial resolution. Targeting the BAV class, Grad-CAM activation volumes are upsampled to input size, min--max normalized, and alpha-blended with the original frames. These base-level explanations are later summarized qualitatively in Section~\ref{sec:results}.

\paragraph{Meta-level SHAP}\label{sec:shap}
The meta layer consumes 10-D features per sample: $[p_{\text{TAV}},p_{\text{BAV}}]$ from each of the five bases. We quantify feature contributions using SHAP~\citep{lundberg2017shap}: TreeExplainer is used for the XGBoost meta-model, while KernelExplainer (model-agnostic) is used for LR, MLP, SVM, the Transformer, and the final deployed ensemble. To avoid leakage, the KernelExplainer background set is formed by a small stratified subsample of $X_{\text{oof}}$ (never from outer-test), and attributions are computed only on $X_{\text{test}}$. For the final ensemble, SHAP was applied directly to the deployed prediction function, which averages the five meta-model BAV probabilities and then applies the fitted recalibration function when enabled.

\section{Results}
\label{sec:results}

\subsection{Overall performance}
Table~\ref{tab:seedwise_results} reports outer-test performance of the calibrated multi-backbone ensemble across 10 independent random seeds under fixed 3-fold stratified outer cross-validation. Point metrics (Accuracy, Precision, Recall, F1) were computed by thresholding the calibrated BAV probability using a fold-specific threshold $\tau^\star$ selected on calibrated meta-OOF outer-train scores (Section~\ref{sec:threshold}). Unless otherwise stated, Precision, Recall, and F1 reported in this section refer to the BAV-positive class.

\begin{table*}[t]
\centering\footnotesize
\caption{Per-seed outer-test performance of the calibrated multi-backbone ensemble (positive class = BAV)
under fixed 3-fold stratified outer cross-validation splits. The final row reports the overall mean~$\pm$~SD.}
\label{tab:seedwise_results}
\begin{tabular}{cccccccc}
\toprule
Seed & Accuracy & Precision & Recall & F1 & AUROC & AP & Brier \\
\midrule
1  & 0.896 & 0.943 & 0.879 & 0.909 & 0.848 & 0.884 & 0.148 \\
2  & 0.892 & 0.940 & 0.875 & 0.905 & 0.846 & 0.881 & 0.150 \\
3  & 0.897 & 0.944 & 0.880 & 0.911 & 0.851 & 0.888 & 0.147 \\
4  & 0.893 & 0.939 & 0.876 & 0.906 & 0.849 & 0.886 & 0.149 \\
5  & 0.890 & 0.938 & 0.873 & 0.903 & 0.847 & 0.883 & 0.150 \\
6  & 0.894 & 0.941 & 0.877 & 0.908 & 0.850 & 0.885 & 0.149 \\
7  & 0.898 & 0.945 & 0.881 & 0.912 & 0.852 & 0.889 & 0.148 \\
8  & 0.891 & 0.939 & 0.874 & 0.904 & 0.848 & 0.884 & 0.151 \\
9  & 0.895 & 0.942 & 0.878 & 0.908 & 0.849 & 0.887 & 0.149 \\
10 & 0.893 & 0.940 & 0.876 & 0.906 & 0.850 & 0.886 & 0.150 \\
\midrule
\textbf{Mean $\pm$ SD} & \textbf{0.894 $\pm$ 0.003} & \textbf{0.941 $\pm$ 0.002} & \textbf{0.877 $\pm$ 0.003} & \textbf{0.907 $\pm$ 0.003} & \textbf{0.849 $\pm$ 0.002} & \textbf{0.885 $\pm$ 0.002} & \textbf{0.149 $\pm$ 0.001} \\
\bottomrule
\end{tabular}
\end{table*}

Across seeds, the ensemble achieved mean outer-test Accuracy $=0.894\pm0.003$, Precision $=0.941\pm0.002$, Recall $=0.877\pm0.003$, and F1-score $=0.907\pm0.003$. Threshold-free discrimination was also strong (AUROC $=0.849\pm0.002$, AP $=0.885\pm0.002$), and probability reliability was reflected by a mean Brier score of $0.149\pm0.001$. Variability across seeds was small (F1 SD $\approx 0.003$), indicating stable optimization under fixed data splits. Note that this SD captures optimization randomness (initialization and training stochasticity), not sampling uncertainty over different train/test partitions. When multiple thresholds tied for maximum F1 on the tuning grid, we selected the smallest $\tau$ to favor Recall, consistent with a screening-oriented objective. Fold-wise thresholds (mean~$\pm$~SD across 10 seeds) and the corresponding tuning-set operating characteristics are reported in Table~\ref{tab:fold_thresholds}.

\begin{table}[!bp]
\centering
\small
\caption{Selected decision thresholds $\tau^\star$ per outer fold. Values are mean $\pm$ SD across 10 random seeds.}
\label{tab:fold_thresholds}
\begin{tabular*}{\columnwidth}{@{\extracolsep{\fill}}llc@{}}
\toprule
Outer fold & Measure & Mean $\pm$ SD \\
\midrule
Fold 1 & $\tau^\star$        & $0.190 \pm 0.132$ \\
       \addlinespace[0.35em]
       & Tune Precision     & $0.877 \pm 0.068$ \\
       \addlinespace[0.35em]
       & Tune Recall        & $0.929 \pm 0.084$ \\
       \addlinespace[0.35em]
       & Tune F1-score      & $0.901 \pm 0.069$ \\
\midrule
Fold 2 & $\tau^\star$        & $0.326 \pm 0.267$ \\
\addlinespace[0.35em]
       & Tune Precision     & $0.868 \pm 0.130$ \\
       \addlinespace[0.35em]
       & Tune Recall        & $0.917 \pm 0.087$ \\
       \addlinespace[0.35em]
       & Tune F1-score      & $0.887 \pm 0.099$ \\
\midrule
Fold 3 & $\tau^\star$        & $0.427 \pm 0.170$ \\
\addlinespace[0.35em]
       & Tune Precision     & $0.812 \pm 0.143$ \\
       \addlinespace[0.35em]
       & Tune Recall        & $0.886 \pm 0.065$ \\
       \addlinespace[0.35em]
       & Tune F1-score      & $0.837 \pm 0.063$ \\
\bottomrule
\end{tabular*}
\end{table}

We emphasize that $\tau^\star$ affects only threshold-dependent point metrics (Accuracy, Precision, Recall, F1) by changing the confusion matrix. Threshold-free metrics (AUROC, AP) and probability-based calibration metrics (Brier score, ECE) are computed directly from calibrated probabilities and do not depend on $\tau^\star$. For global ROC/PR, calibration, and decision-curve visualization, we use a single seed-averaged calibrated probability per study, $\bar p_i=\frac{1}{S}\sum_{s=1}^{S} p_i^{(s)}$ with $S=10$ (fixed outer splits), so each patient contributes one prediction.

\subsection{ROC and Precision--Recall curves}
\begin{figure*}[!tbp]
  \centering
  \begin{subfigure}[t]{0.44\textwidth}
    \centering
    \includegraphics[width=\linewidth]{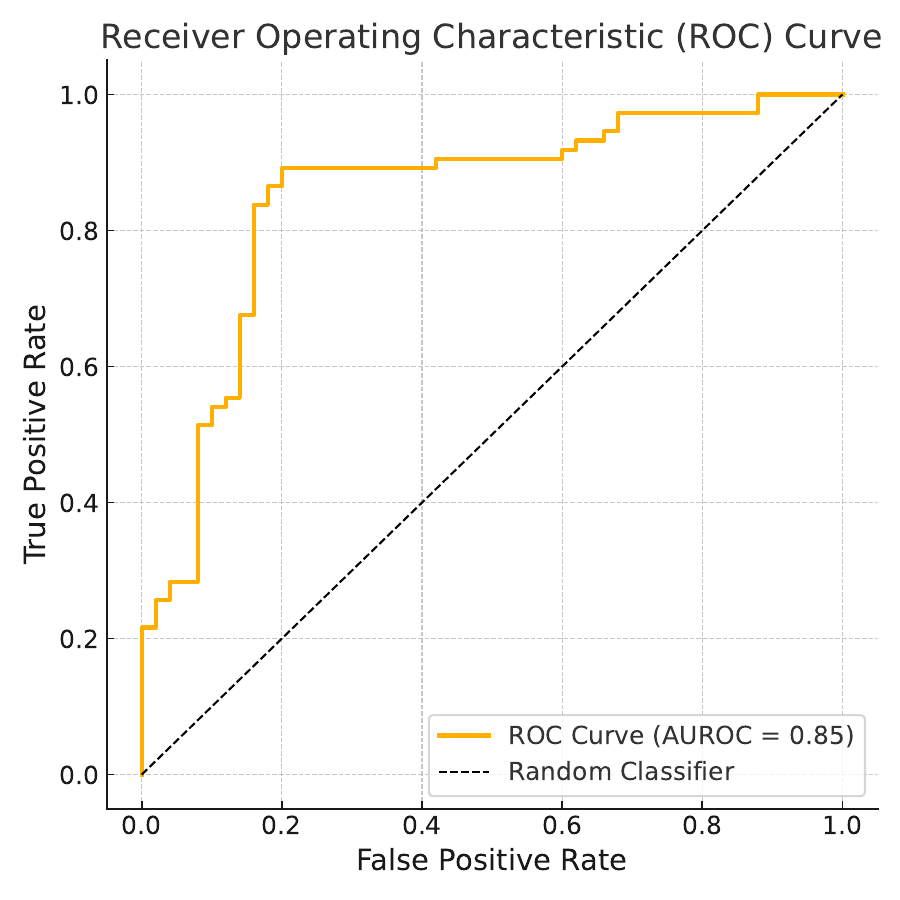}
    \subcaption{\textbf{Global ROC.} AUROC = 0.849.}
    \label{fig:roc_global}
  \end{subfigure}\hfill
  \begin{subfigure}[t]{0.44\textwidth}
    \centering
    \includegraphics[width=\linewidth]{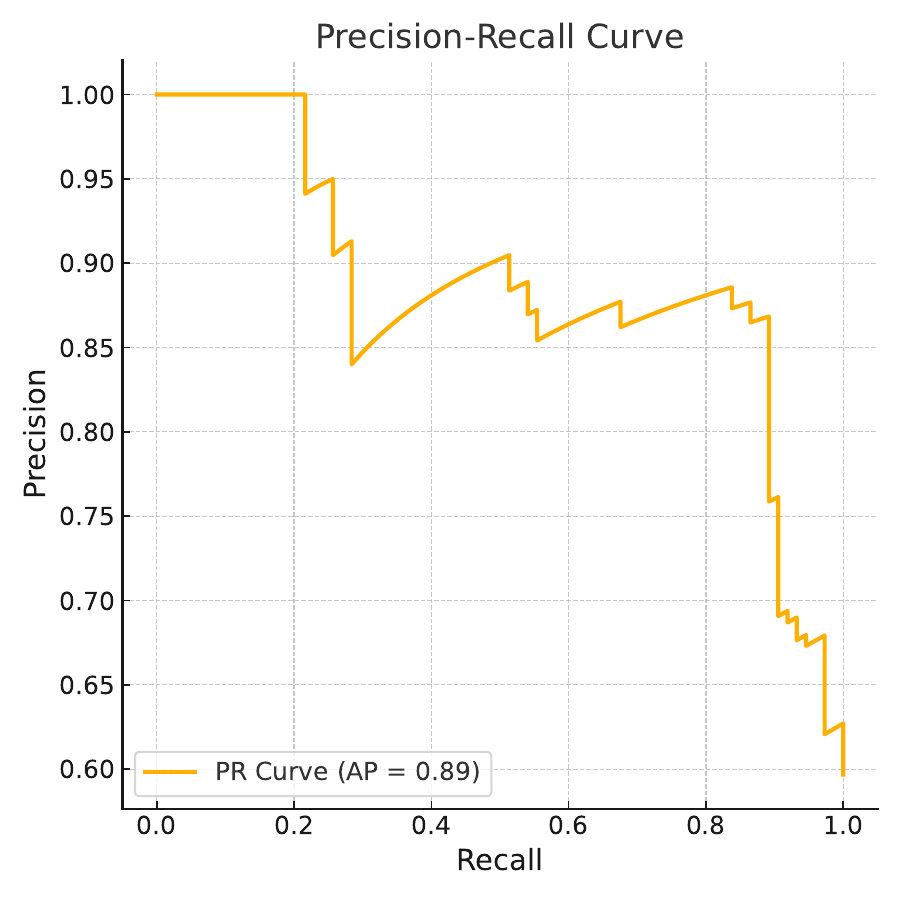}
    \subcaption{\textbf{Global PR.} AP = 0.885.}
    \label{fig:pr_global}
  \end{subfigure}
  \caption{\textbf{Global performance curves across all outer-test samples.} Curves are computed using the seed-averaged calibrated probabilities $\bar p_i$ (10 seeds; fixed outer splits), so each patient contributes a single point.}
  \label{fig:roc_pr_global}
\end{figure*}

Figure~\ref{fig:roc_pr_global} summarizes threshold-free performance using ROC and Precision--Recall curves. For these global visualizations we use one calibrated probability per study, obtained by seed-averaging the calibrated predictions across the 10 random seeds under fixed outer splits ($\bar p_i=\frac{1}{S}\sum_{s=1}^{S} p_i^{(s)}$, $S{=}10$). This avoids pseudo-replication (each patient contributes a single point) while keeping the per-seed mean $\pm$ SD summary in Table~\ref{tab:seedwise_results} unchanged.

The global ROC in Fig.~\ref{fig:roc_global} demonstrates strong discrimination (AUROC = 0.849), consistent with the
seed-wise summary in Table~\ref{tab:seedwise_results}. The corresponding PR curve in Fig.~\ref{fig:pr_global}
achieves AP = 0.885 and highlights high precision in the clinically relevant mid- to high-recall regime.

\subsection{Stability across random seeds}
Across the 10 seeds, performance was tightly clustered with no seed exhibiting catastrophic degradation. F1-scores remained near 0.90, AUROC stayed in the 0.84--0.86 range, and AP in approximately 0.87--0.90. This narrow spread suggests the OOF stacking pipeline is robust to initialization and training stochasticity under fixed outer splits (Table~\ref{tab:seedwise_results}).

\subsection{Calibration and operating characteristics}
\begin{figure}[t]
  \centering
  \includegraphics[width=\linewidth]{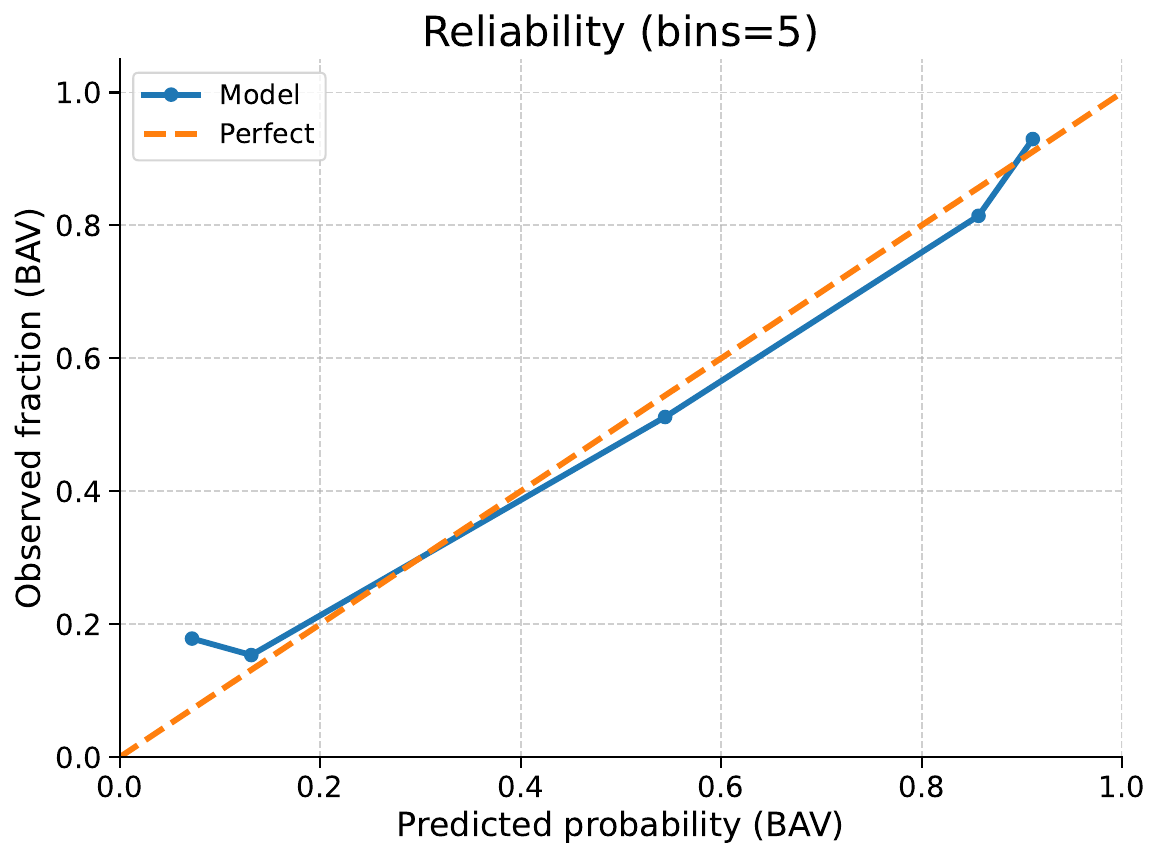}
  \caption{\textbf{Global reliability diagram (5 bins).} Observed fraction of BAV versus predicted BAV probability for the calibrated ensemble on the held-out outer-test folds. The plot is computed using the seed-averaged calibrated probabilities $\bar p_i$. The dashed line denotes perfect calibration.}
  \label{fig:calibration_curve}
\end{figure}

Post-hoc Platt scaling was applied at the meta level to improve the reliability of the ensemble's output
probabilities. Calibration on the held-out outer-test folds is summarized by a mean Brier score of $0.149\pm0.001$
and the global reliability diagram in Fig.~\ref{fig:calibration_curve}. For global visualization, we use the seed-averaged calibrated probabilities $\bar p_i$ defined above, ensuring that each patient contributes a single prediction. Because operating thresholds and decision-curve analysis are defined on this absolute probability scale, calibrated probabilities are important so that a chosen threshold corresponds to a consistent risk level across folds and random seeds. Clinical utility was further examined using decision curve analysis (Section~\ref{sec:dca}). Expected calibration error (ECE; 5 bins) is computed from the same predictions. Across the three outer-test folds, the expected calibration error (ECE; 5 bins) was $0.151 \pm 0.041$
(fold-wise: 0.105, 0.183, 0.166), consistent with Fig.~\ref{fig:calibration_curve}.

\subsection{Decision curve analysis}\label{sec:dca}
\begin{figure}[t]
  \centering
  \includegraphics[width=\columnwidth]{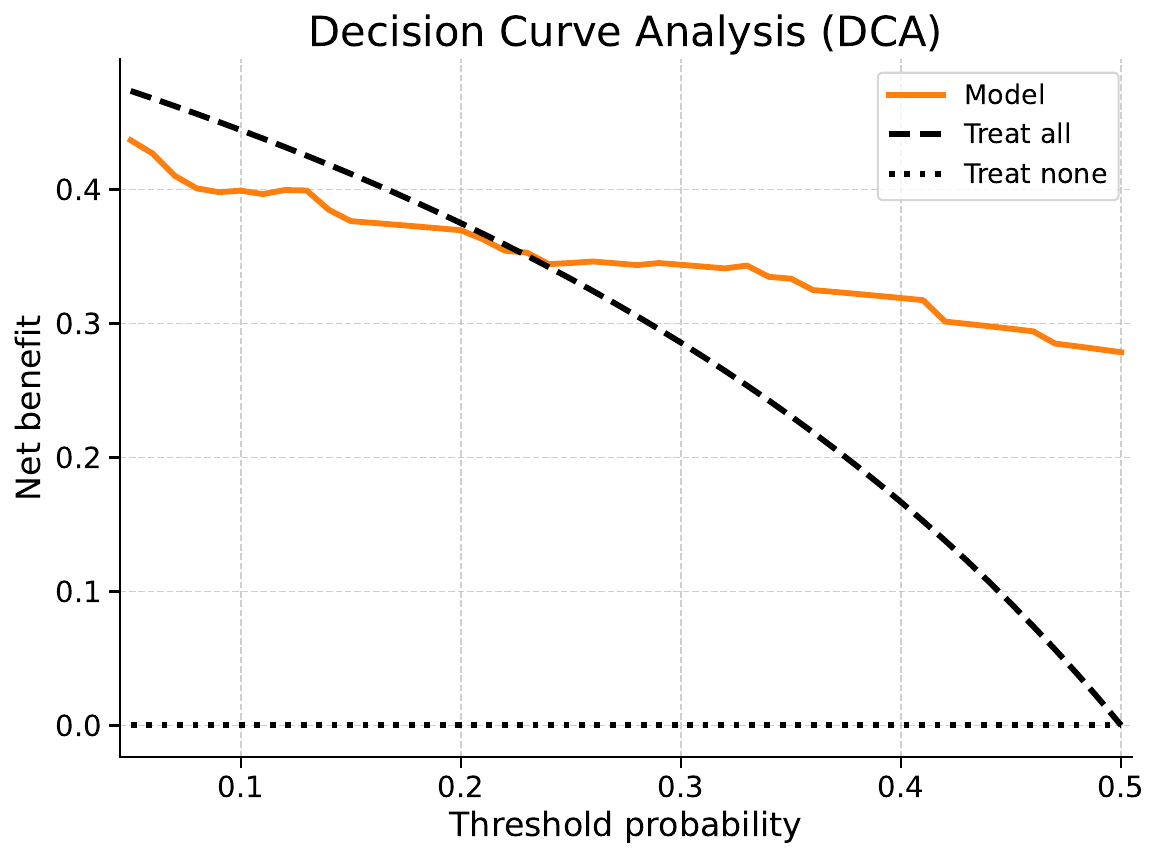}
  \caption{\textbf{Decision curve analysis (global).} Net benefit as a function of threshold probability for the calibrated ensemble compared with treat-all and treat-none strategies. The curve is computed using the seed-averaged calibrated probabilities $\bar p_i$ (10 seeds; fixed outer splits).}
  \label{fig:dca_global}
\end{figure}

We assessed the potential clinical utility using Decision Curve Analysis (DCA)~\citep{vickers2006dca}, which reports the net benefit as a function of the threshold probability $p_t$ by comparing the model to \emph{treat-all} and \emph{treat-none} strategies. The net benefit was computed as
\[
\mathrm{NB}(p_t)=\frac{\mathrm{TP}}{N}-\frac{\mathrm{FP}}{N}\cdot \frac{p_t}{1-p_t},
\]
with $\mathrm{NB}=0$ for treat-none and the standard prevalence-based formulation for treat-all. Figure~\ref{fig:dca_global} shows the global DCA curve constructed using the seed-averaged calibrated
probabilities $\bar p_i$, so each patient contributes a single prediction to the net benefit calculation.

Across the examined range, the model achieves a higher net benefit than treat-none for all threshold probabilities. Compared to treat-all, the model provides a higher net benefit for moderate-to-high thresholds (approximately $p_t \gtrsim 0.23$), while treat-all is slightly preferred at lower thresholds, consistent with the relatively high  prevalence of BAV in our cohort.

\subsection{Explainability findings}

\subsubsection{Frame-level Grad-CAM}
\begin{figure}[!th]
  \centering
  \captionsetup[sub]{justification=centering}
  \begin{subfigure}[t]{\linewidth}
    \includegraphics[width=\linewidth]{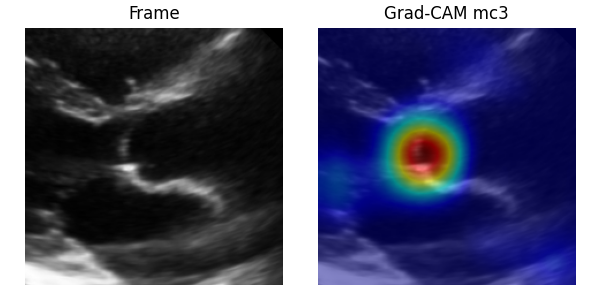}
    \caption{MC3}
  \end{subfigure}\par\vspace{6pt}
  \begin{subfigure}[t]{\linewidth}
    \includegraphics[width=\linewidth]{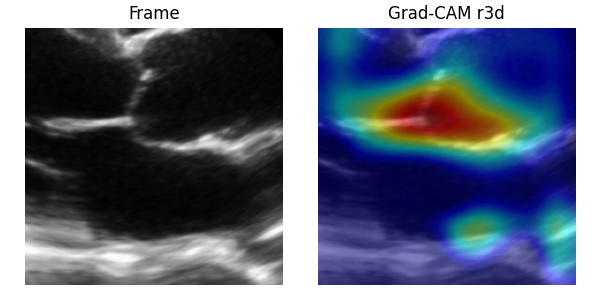}
    \caption{R3D}
  \end{subfigure}\par\vspace{6pt}
  \begin{subfigure}[t]{\linewidth}
    \includegraphics[width=\linewidth]{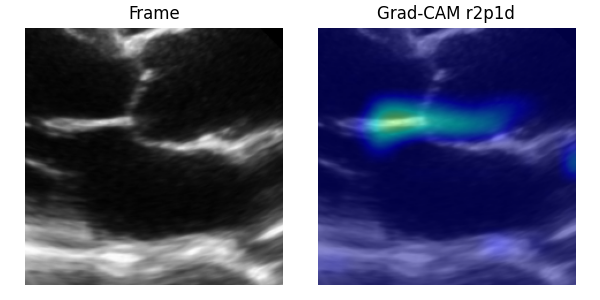}
    \caption{R2P1D}
  \end{subfigure}
  \caption{\textbf{Base-level Grad-CAM overlays.} Representative examples from three representative backbones (MC3, R3D, and R2P1D) in one seed and outer fold; X3D and S3D showed qualitatively similar localization patterns and are not shown for brevity. Each panel shows a raw PLAX frame (left) and the Grad-CAM overlay for the BAV class (right) computed from the final fine-tuned model of the corresponding backbone.}
  \label{fig:gradcam_bases}
\end{figure}

Figure~\ref{fig:gradcam_bases} shows representative Grad-CAM overlays for three representative backbones (MC3, R3D, and R2P1D) from a single seed and outer fold; X3D and S3D exhibited qualitatively similar localization patterns, but are omitted for brevity. Saliency consistently concentrates around the aortic root and leaflet coaptation region, particularly during early systole when commissural geometry and doming are most apparent. Correct classifications display compact and physiologically coherent activation, whereas misclassifications exhibit diffuse or off-target patterns, typically due to poor acoustic windows, shadowing, or off-axis PLAX acquisition.

\subsubsection{Meta-level SHAP}
\begin{figure}[!tbp]
  \centering
  \includegraphics[width=\linewidth]{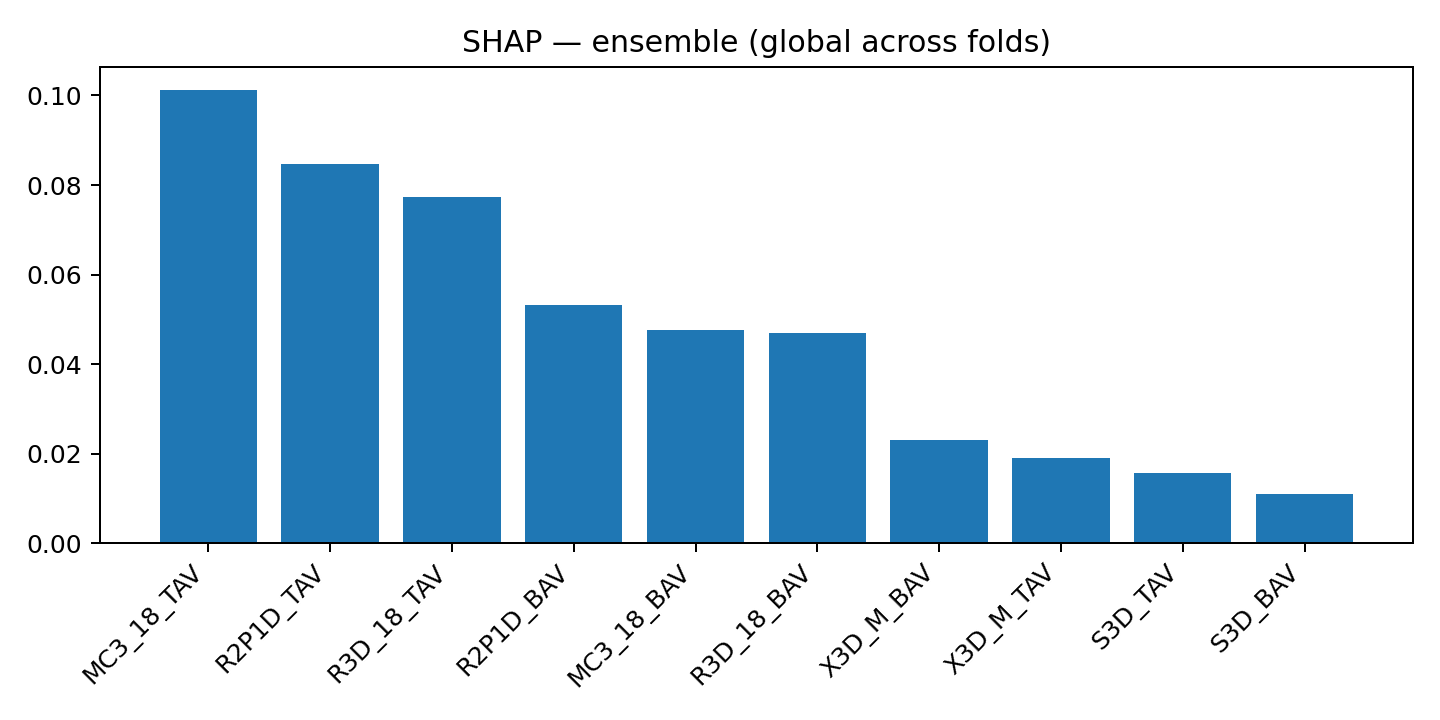}
  \caption{\textbf{Global SHAP importance for the final deployed ensemble.} Bars show mean absolute SHAP values for the 10 meta-features, computed by applying KernelExplainer directly to the final ensemble prediction function that averages the five meta-model BAV probabilities (with recalibration applied when enabled), and then averaging absolute SHAP values across the explained outer-test samples. The ensemble relied most strongly on TAV probabilities (negative predictors) from MC3, R2P1D, and R3D, while BAV probabilities contributed secondary positive evidence. S3D features contributed least overall.}
  \label{fig:shap_bar}
\end{figure}

To interpret the stacked ensemble, we computed SHAP values on the 10-dimensional meta-features ($[p_{\text{TAV}}, p_{\text{BAV}}]$ from each backbone). Figure~\ref{fig:shap_bar} reports mean absolute SHAP values for the final deployed ensemble, obtained by applying KernelExplainer directly to the ensemble prediction function that averages the five meta-model BAV probabilities (with recalibration applied when enabled), and then averaging absolute SHAP values across the explained outer-test samples. The largest contributions came from the \emph{TAV probabilities} of MC3, R2P1D, and R3D (strong negative evidence for BAV), while BAV probabilities provided smaller positive contributions; S3D features were least influential.

Overall, the meta-learner appears to identify BAV primarily by \emph{discounting} high-confidence tricuspid morphology, while integrating supportive BAV cues when present. We observed the same qualitative SHAP ranking and directionality across folds and across the evaluated meta-learners.

\subsection{Error analysis}
Misclassifications were most commonly observed in cases with challenging image quality or ambiguous valve morphology. False negatives often involved BAV cases with very mild systolic asymmetry, whereas false positives tended to occur in TAV cases with eccentric or domed opening patterns that can resemble partial cusp fusion in PLAX imaging.

\begin{figure}[!tb]
  \centering
  \captionsetup[sub]{justification=centering}

  \begin{subfigure}[t]{\linewidth}
    \includegraphics[width=\linewidth]{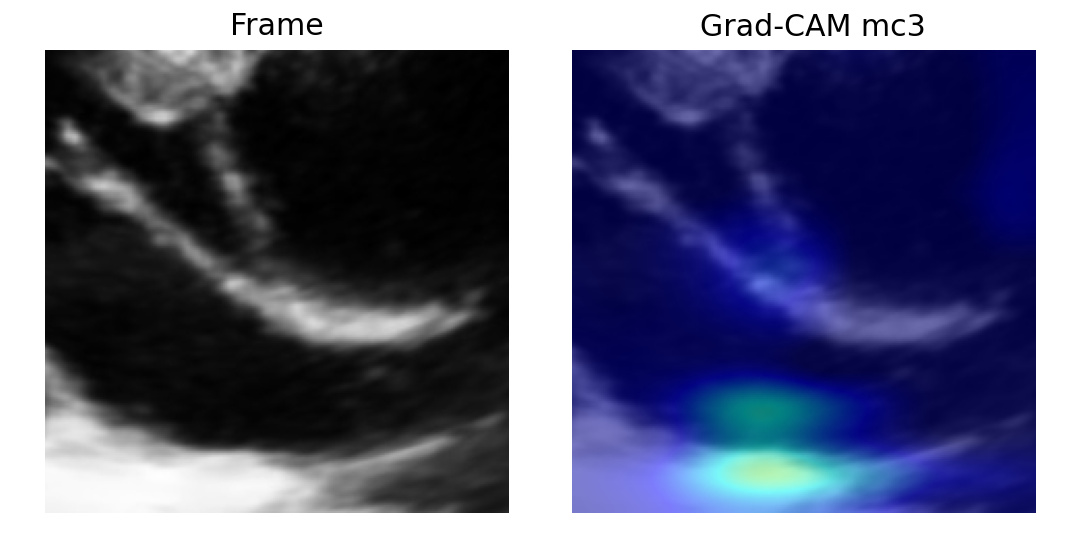}
    \caption{MC3 misclassification}
  \end{subfigure}
  \par\vspace{6pt}

  \begin{subfigure}[t]{\linewidth}
    \includegraphics[width=\linewidth]{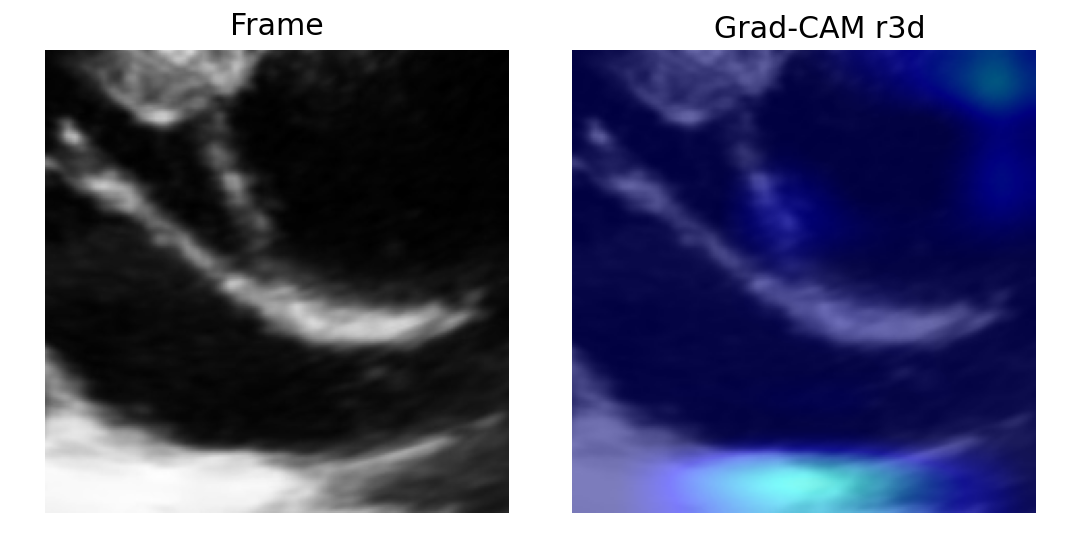}
    \caption{R3D misclassification}
  \end{subfigure}
  \par\vspace{6pt}

  \begin{subfigure}[t]{\linewidth}
    \includegraphics[width=\linewidth]{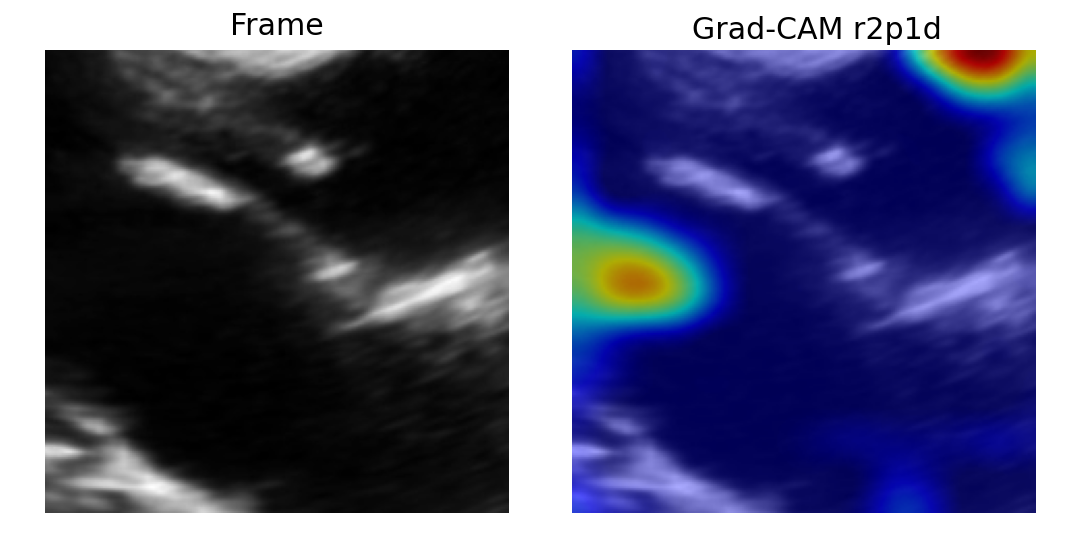}
    \caption{R2P1D misclassification}
  \end{subfigure}

  \caption{\textbf{Representative misclassified cases with Grad-CAM overlays.} Each panel shows a raw PLAX frame (left) and its Grad-CAM map (right). Saliency is often displaced from the aortic valve region, potentially due to off-axis imaging, artifacts, or low SNR. These patterns varied by case but were stable across seeds, indicating that errors likely arise from data limitations rather
  than model instability.}
  \label{fig:gradcam_misclassified}
\end{figure}

To explore these errors, Fig.~\ref{fig:gradcam_misclassified} presents representative Grad-CAM overlays for three misclassified cases across different backbones (R2P1D, R3D, MC3). In these examples, the saliency maps appear displaced away from the aortic valve and highlight regions that are not obviously related to leaflet motion. While Grad-CAM does not provide a definitive causal explanation for model behaviour, several plausible factors may contribute to these patterns:

\begin{itemize}
\item \textbf{Off-axis PLAX views} may cause the valve to be partially visible or poorly centered, leading the network to rely on alternative structures with stronger gradients.
\item \textbf{Shadowing and reverberation artifacts} can introduce high-contrast edges that Grad-CAM may interpret as influential, even if they are anatomically irrelevant.
\item \textbf{Low signal-to-noise frames} may reduce visibility of commissural geometry, causing the model to attend to more stable but non-diagnostic regions.
\end{itemize}

Importantly, these mislocalizations were consistent across seeds for the same cases, suggesting that they reflect limitations of the input data or the interpretability method rather than instability of the learning algorithm. Improving acquisition quality, incorporating view-quality assessment, or using more robust attribution methods may help mitigate these error modes in future work.

\subsection{Summary}
The leakage-aware, calibrated multi-backbone ensemble achieved strong and highly reproducible performance for BAV/TAV classification from routine PLAX cine loops, with a mean F1-score of $0.907$ and AUROC of $0.849$ across 10 independent seeds. Probability calibration produced well-behaved and interpretable risk estimates, and the combination of base-level Grad-CAM and meta-level SHAP enabled transparent auditing of both spatial and feature-level model behaviour. Performance variability across seeds was minimal, indicating that the nested OOF stacking pipeline is stable and not dependent on favorable initialization. Taken together, these findings suggest that the proposed framework is technically robust and potentially suitable for downstream clinical workflows, particularly in non-specialist or heterogeneous echocardiographic acquisition settings.

\section{Discussion}
\label{sec:discussion}

\subsection{Principal findings}
This study demonstrates that a leakage-aware multi-backbone video ensemble can classify bicuspid versus tricuspid aortic valves directly from routine PLAX cine loops with robust and well-calibrated performance (F1 = 0.907 $\pm$ 0.003; AUROC = 0.849 $\pm$ 0.002). These results indicate that diagnostically informative spatio--temporal cues are present even in non-specialist PLAX acquisitions, supporting the feasibility of PLAX-based automated pre-diagnostic screening. The integration of multiple complementary video backbones with strict out-of-fold (OOF) stacking contributed to stable generalization despite the modest dataset size, while dual-level explainability (Grad-CAM and SHAP) provided transparency at both spatial and probabilistic levels.

\subsection{Comparison with prior work}
Prior work on automated valve morphology assessment has predominantly focused on the parasternal short-axis (PSAX) view or on frame-based CNNs that do not model temporal information. In contrast, our approach leverages motion-aware 3D-CNN architectures trained on PLAX cine loops, a view that is easier to acquire and more frequently available in routine practice. Previous still-frame or PSAX-based approaches have reported AUROC values around 0.75--0.80 (e.g.~\citep{chen2020echocnn}); our ensemble achieved an AUROC of 0.85 while also improving calibration and explainability. Methodologically, the introduced leakage-aware stacking protocol addresses a common but underrecognized pitfall in small-cohort medical imaging studies: inadvertent meta-level leakage during ensemble training. By ensuring that all meta-features originate strictly from out-of-sample predictions, the proposed framework yields more reliable estimates of generalization than conventional stacking or naive cross-validation.

\subsection{Clinical and methodological implications}
These findings suggest that PLAX-based cine analysis may increase non-specialist screening by providing rapid, reproducible, and explainable assessments of valve morphology using routinely acquired echocardiographic views. This is particularly relevant for primary care and community settings, where acquisition expertise and availability of PSAX imaging may be limited. From a methodological standpoint, the study highlights the utility of carefully designed stacking ensembles for small datasets: diverse pretrained video encoders capture complementary spatio--temporal cues, while meta-level fusion and post-hoc calibration support reliable and interpretable probability estimates. Together, these components align with contemporary recommendations for reliable and auditable clinical AI.

An additional methodological observation concerns input normalization. Although the video backbones were pretrained on ImageNet/Kinetics and expect mean--std normalized RGB inputs, preliminary experiments showed that applying these normalization statistics to grayscale PLAX frames degraded performance. This likely reflects substantial distributional differences: natural RGB images span wide intensity ranges and color channels, whereas B-mode ultrasound is single-channel, speckle-dominated, and concentrated within a narrow dynamic range. Imposing RGB-style normalization can therefore shift ultrasound textures into activation regimes mismatched to pretrained filters. Training directly on $[0,1]$--scaled grayscale inputs preserved native ultrasound statistics and produced more stable optimization and better outer-CV performance. This underscores the importance of modality-specific preprocessing when adapting pretrained video models to medical imaging.

\paragraph{Augmentation choice and flipping.}
Horizontal flipping is a standard augmentation in natural video recognition, where left--right mirroring is typically label-preserving and encourages viewpoint invariance. In transthoracic echocardiography, however, PLAX cine loops are commonly acquired and displayed under a standardized orientation, and mirroring can create anatomically implausible presentations and disrupt learned spatial priors. In our preliminary ablation, enabling horizontal flipping during training did not improve and in fact degraded performance and/or probability reliability. Vertical flipping was also avoided, as it can invert the expected anatomical orientation and further disrupt spatial priors. We therefore retained only conservative spatial perturbations (small rotations and translations), which preserve morphology cues while improving robustness to minor probe/view variability.

\subsection{Limitations}
Several limitations should be acknowledged. First, the dataset was modest ($N{=}90$) and derived entirely from a single clinical site, which may limit generalizability across different scanner vendors, acquisition protocols, and patient populations. Only grayscale PLAX B-mode loops were analyzed, excluding PSAX views and Doppler modalities that could further enhance diagnostic discrimination. Second, the training pipeline was implemented in a \emph{centralized} manner, requiring transfer of patient-derived cine loops for model development. Even with anonymization, such data movement may introduce privacy and governance challenges under strict regulatory frameworks. Third, although OOF stacking mitigates meta-level leakage, the limited OOF sample size restricts the complexity of meta-learners and may underrepresent predictive uncertainty in borderline cases. Finally, explainability analyses relied on qualitative Grad-CAM and SHAP evaluation; formal clinical reader studies are required to assess whether model saliency aligns with expert diagnostic reasoning.

In addition, fold-wise threshold selection showed noticeable variability across seeds (Table~\ref{tab:fold_thresholds}) in this small cohort, and one fold required a more conservative operating point. While this procedure avoids test leakage, it highlights that the chosen precision--recall trade-off can be sensitive to limited tuning data; future work will evaluate clinically fixed thresholds or cost-sensitive objectives on larger multi-center cohorts.

\subsection{Future work}
\label{sec:future_work}
Future research should focus on validating the proposed framework in larger, multi-center cohorts with broader scanner, protocol, and patient variability. External validation is essential to determine whether the observed PLAX-based performance generalizes beyond the present single-center dataset. Incorporating additional echocardiographic views, particularly PSAX, as well as Doppler information, may further improve sensitivity and robustness in challenging cases.

A natural extension is to transition the pipeline from centralized training to federated learning. In a federated setting, multiple hospitals could collaboratively train or fine-tune the model without sharing raw echocardiographic cine loops, reducing privacy and data-governance barriers. Prior work has explored client selection and imbalance-aware federated learning~\citep{NikolaidisEtAl2023FMEC_DropoutFL,NikolaidisEfraimidis2025Cluster}, explainability in distributed medical pipelines~\citep{BriolaEtAl2024EICC_FedXAI_BreastCancer}, membership-inference risks in federated healthcare~\citep{NikolaidisEfraimidis2025ComputingMIA}, and federated anomaly detection for screening~\citep{PavlidisEtAl2024Arxiv_FedAnomaly_ASD}. The proposed leakage-aware evaluation design, per-sample logging, calibration, and explainability components are well suited to such distributed settings, where auditability and transparent model behavior are especially important.

Beyond algorithmic refinement, prospective evaluation in real-world clinical workflows and clinician-in-the-loop studies will be essential to assess usability, diagnostic reliability, and trust. Ultimately, federated, explainable, and well-calibrated PLAX-based pipelines may form a foundation for privacy-preserving AI systems in cardiovascular imaging.

\section{Conclusion}
\label{sec:conclusion}

This work introduced a leakage-aware, multi-backbone video ensemble for automated classification of bicuspid versus tricuspid aortic valves from routine PLAX echocardiographic cine loops. The proposed framework combines diverse pretrained video encoders with out-of-fold stacking, meta-level probability calibration, and dual-level explainability, achieving robust and well-calibrated performance on a small, single-center dataset. Across 10 independent seeds with fixed outer splits, the ensemble attained mean outer-test performance of Accuracy $= 0.894 \pm 0.003$, F1-score $= 0.907 \pm 0.003$, AUROC $= 0.849 \pm 0.002$, and AP $= 0.885 \pm 0.002$, with a calibrated Brier score of $0.149 \pm 0.001$. Grad-CAM and SHAP analyses provided complementary spatial and probabilistic transparency, supporting case-level auditing and helping to identify acquisition-related failure modes.

The findings demonstrate that diagnostically relevant motion cues in standard PLAX views can support reliable, interpretable valve morphology classification, even in non-specialist acquisition settings. This suggests that PLAX-based video ensembles could assist non-expert cardiologists in population screening for BAV, enabling earlier specialist referral and reducing the risk of missed diagnosis with subsequent major valvular and aortic complications.

Future work will focus on expanding the dataset to multi-center cohorts and transitioning the pipeline toward federated learning to enable privacy-preserving, cross-institutional training without sharing raw data. Integrating multi-view echocardiographic inputs and conducting clinician-in-the-loop, prospective validation studies will be essential to further enhance generalizability, interpretability, and real-world applicability. In summary, the proposed approach establishes a reproducible foundation for trustworthy, explainable AI in echocardiographic valve assessment.

\section*{Ethics and data governance}
This retrospective study used de-identified transthoracic echocardiography cine loops exported in DICOM format from routine clinical care at a single center, under institutional oversight and in accordance with the Declaration of Helsinki and applicable local regulations. All direct identifiers were removed prior to research access; only grayscale PLAX B-mode cine data and minimal metadata required for modeling were retained. Patients provided informed consent for the use of anonymized data for research purposes.

Model outputs are intended as calibrated decision-support scores rather than definitive diagnoses and should be interpreted by qualified clinicians. The dataset cannot be shared publicly due to patient privacy; code and trained weights may be made available upon reasonable request, subject to institutional approval.

\bibliographystyle{unsrtnat}

\bibliography{cas-refs}

@article{chen2020echocnn,
  title={Automatic classification of {BAV} using {CNN}s on echocardiography},
  author={Chen, J. and Li, S. and Xiao, Y.},
  journal={IEEE Journal of Biomedical and Health Informatics},
  volume={24},
  number={11},
  pages={3120--3130},
  year={2020},
  publisher={IEEE}
}

@article{ouyang2020video,
  title={Video-based {AI} for beat-to-beat assessment of cardiac function},
  author={Ouyang, D. and He, T. and Cao, C. and others},
  journal={Nature},
  volume={580},
  number={7802},
  pages={252--256},
  year={2020},
  publisher={Nature Publishing Group}
}

@article{hong2022hypertrophic,
  title={Transformer-based deep learning for hypertrophic cardiomyopathy diagnosis from echocardiographic videos},
  author={Hong, Y. and Li, X. and Tang, Y.},
  journal={Computer Methods and Programs in Biomedicine},
  volume={221},
  pages={106895},
  year={2022},
  publisher={Elsevier}
}

@article{hillebrand2017accuracy,
  title={Accuracy of transthoracic echocardiography to diagnose bicuspid aortic valve: comparison with surgical findings and cardiac magnetic resonance imaging},
  author={Hillebrand, M. and Li, W. and Gross, S. and others},
  journal={International Journal of Cardiology},
  volume={243},
  pages={260--265},
  year={2017},
  publisher={Elsevier}
}

@article{ayad2011echoaccuracy,
  title={Echocardiographic accuracy in diagnosing bicuspid aortic valve and its clinical implications},
  author={Ayad, R.F. and Puthumana, J.J. and Doukky, R.},
  journal={Echocardiography},
  volume={28},
  number={5},
  pages={558--562},
  year={2011},
  publisher={Wiley}
}

@article{howard2020improving,
  title={Improving ultrasound video classification: an evaluation of novel {CNN} architectures},
  author={Howard, James P. and Tan, Jeremy and Shun-Shin, Matthew J. and Mahdi, Dina and Nowbar, Alexandra N. and Arnold, Ahran D. and Ahmad, Yousif and McCartney, Peter and Zolgharni, Massoud and Linton, Nick W. F. and Sutaria, Nilesh and Rana, Bushra and Mayet, Jamil and Rueckert, Daniel and Cole, Graham D. and Francis, Darrel P.},
  journal={Journal of Medical Artificial Intelligence},
  volume={3},
  number={4},
  pages={e200003},
  year={2020},
  doi={10.21037/jmai.2019.10.03}
}

@article{muller2022ensemble,
  title={An analysis on ensemble learning optimized medical image classification with deep convolutional neural networks},
  author={M{\"u}ller, Dominik and Soto-Rey, I{\~n}aki and Kramer, Frank},
  journal={IEEE Access},
  volume={10},
  pages={66467--66480},
  year={2022},
  doi={10.1109/ACCESS.2022.3182399}
}

@article{ganaie2022ensembleReview,
  title={Ensemble deep learning: A review},
  author={Ganaie, M. A. and Hu, Minghui and Malik, A. K. and Tanveer, M. and Suganthan, P. N.},
  journal={Engineering Applications of Artificial Intelligence},
  volume={115},
  pages={105151},
  year={2022},
  doi={10.1016/j.engappai.2022.105151}
}

@article{ganie2025heartdisease,
  title={Ensemble learning with explainable {AI} for improved heart disease prediction},
  author={Ganie, Shahid Mohammad and Pramanik, Pijush Kanti Dutta and Zhao, Zhongming},
  journal={Scientific Reports},
  volume={15},
  pages={13912},
  year={2025},
  doi={10.1038/s41598-025-97547-6}
}

@inproceedings{selvaraju2017gradcam,
  title={Grad-{CAM}: Visual explanations from deep networks via gradient-based localization},
  author={Selvaraju, Ramprasaath R. and Cogswell, Michael and Das, Abhishek and Vedantam, Ramakrishna and Parikh, Devi and Batra, Dhruv},
  booktitle={Proceedings of the IEEE International Conference on Computer Vision (ICCV)},
  year={2017},
  pages={618--626},
  doi={10.1109/ICCV.2017.74}
}

@article{lundberg2017shap,
  title={A unified approach to interpreting model predictions},
  author={Lundberg, Scott M. and Lee, Su-In},
  journal={Advances in Neural Information Processing Systems (NeurIPS)},
  volume={30},
  year={2017}
}

@article{vandervelden2022xaiReview,
  title={Explainable artificial intelligence in medical imaging: A systematic review},
  author={van der Velden, B.H.M. and van de Leemput, S.C. and Veldhuis, W.B. and Gilhuijs, K.G.A.},
  journal={Medical Image Analysis},
  volume={79},
  pages={102470},
  year={2022},
  doi={10.1016/j.media.2022.102470}
}

@article{NikolaidisEfraimidis2025Cluster,
  title   = {Advancing Elderly Social Care Dropout Prediction with Federated Learning: Client Selection and Imbalanced Data Management},
  author  = {Nikolaidis, Christos Chrysanthos and Efraimidis, Pavlos S.},
  journal = {Cluster Computing},
  year    = {2025},
  volume  = {28},
  number  = {2},
  pages   = {114},
  doi     = {10.1007/s10586-024-04850-4}
}

@article{NikolaidisEfraimidis2025ComputingMIA,
  title   = {A Study of Membership Inference Attacks on a Federated Health Care Application},
  author  = {Nikolaidis, Christos Chrysanthos and Efraimidis, Pavlos S.},
  journal = {Computing},
  year    = {2025},
  volume  = {107},
  pages   = {149},
  doi     = {10.1007/s00607-025-01507-x}
}

@inproceedings{BriolaEtAl2024EICC_FedXAI_BreastCancer,
  title     = {A Federated Explainable {AI} Model for Breast Cancer Classification},
  author    = {Briola, Eleni and Nikolaidis, Christos Chrysanthos and Perifanis, Vasileios and Pavlidis, Nikolaos and Efraimidis, Pavlos S.},
  booktitle = {Proceedings of the 2024 European Interdisciplinary Cybersecurity Conference (EICC '24)},
  year      = {2024},
  publisher = {ACM},
  doi       = {10.1145/3655693.3660255}
}

@misc{PavlidisEtAl2024Arxiv_FedAnomaly_ASD,
  title        = {Federated Anomaly Detection for Early-Stage Diagnosis of Autism Spectrum Disorders Using Serious Game Data},
  author       = {Pavlidis, Nikolaos and Perifanis, Vasileios and Briola, Eleni and Nikolaidis, Christos-Chrysanthos and Katsiri, Eleftheria and Efraimidis, Pavlos S. and Filippidou, Despina Elisabeth},
  year         = {2024},
  eprint       = {2410.20003},
  archivePrefix= {arXiv},
  primaryClass = {cs.LG},
  howpublished = {arXiv preprint arXiv:2410.20003},
  url          = {https://arxiv.org/abs/2410.20003}
}

@inproceedings{NikolaidisEtAl2023FMEC_DropoutFL,
  title     = {Federated Learning for Early Dropout Prediction on Healthy Ageing Applications},
  author    = {Nikolaidis, Christos Chrysanthos and Perifanis, Vasileios and Pavlidis, Nikolaos and Efraimidis, Pavlos S.},
  booktitle = {2023 8th International Conference on Fog and Mobile Edge Computing (FMEC)},
  year      = {2023},
  publisher = {IEEE},
  doi       = {10.1109/FMEC59214.2023.10306129}
}

@article{lee2023hybridEnsemble,
  title={Hybrid {CNN}-transformer ensembles for medical video classification},
  author={Lee, S. and Park, D. and Kim, H.},
  journal={Medical Image Analysis},
  year={2023},
  volume={88},
  pages={102819}
}

@article{zhou2024echosmallDL,
  title={Self-supervised echocardiographic representation learning for small datasets},
  author={Zhou, J. and Liang, T. and Wang, C.},
  journal={IEEE Transactions on Medical Imaging},
  year={2024},
  volume={43},
  number={5},
  pages={1223--1235}
}

@article{holste2023severeAS,
  author  = {Holste, Gregory and Oikonomou, Evangelos K. and Mortazavi, Bobak J. and Coppi, Andreas and Faridi, Kamil F. and Miller, Edward J. and Forrest, John K. and McNamara, Robert L. and Ohno-Machado, Lucila and Yuan, Neal and Gupta, Aakriti and Ouyang, David and Krumholz, Harlan M. and Wang, Zhangyang and Khera, Rohan},
  title   = {Severe aortic stenosis detection by deep learning applied to echocardiography},
  journal = {European Heart Journal},
  year    = {2023},
  volume  = {44},
  number  = {43},
  pages   = {4592--4604},
  doi     = {10.1093/eurheartj/ehad456}
}

@incollection{platt1999probabilistic,
  author    = {Platt, John},
  title     = {Probabilistic Outputs for Support Vector Machines and Comparisons to Regularized Likelihood Methods},
  booktitle = {Advances in Large Margin Classifiers},
  publisher = {MIT Press},
  year      = {1999}
}

@article{vickers2006dca,
  author  = {Vickers, Andrew J. and Elkin, Elena B.},
  title   = {Decision curve analysis: a novel method for evaluating prediction models},
  journal = {Medical Decision Making},
  year    = {2006},
  volume  = {26},
  number  = {6},
  pages   = {565--574},
  doi     = {10.1177/0272989X06295361}
}

@article{brier1950verification,
  author  = {Brier, Glenn W.},
  title   = {Verification of Forecasts Expressed in Terms of Probability},
  journal = {Monthly Weather Review},
  year    = {1950},
  volume  = {78},
  number  = {1},
  pages   = {1--3},
  doi     = {10.1175/1520-0493(1950)078<0001:VOFEIT>2.0.CO;2}
}

@incollection{BravermanCheng2021BAV,
  author    = {Braverman, Alan C. and Cheng, Andrew},
  title     = {The Bicuspid Aortic Valve and Associated Aortic Disease},
  booktitle = {Valvular Heart Disease: A Companion to Braunwald's Heart Disease},
  editor    = {Otto, Catherine M. and Bonow, Robert O.},
  edition   = {5},
  publisher = {Saunders/Elsevier},
  year      = {2021},
  pages     = {197--223},
  chapter   = {11},
  isbn      = {9780323546331}
}

@article{Michelena2021BAVConsensusEJCTS,
  author    = {Michelena, Hector I. and Della Corte, Alessandro and Evangelista, Arturo and others},
  title     = {International consensus statement on nomenclature and classification of the congenital bicuspid aortic valve and its aortopathy, for clinical, surgical, interventional and research purposes},
  journal   = {European Journal of Cardio-Thoracic Surgery},
  year      = {2021},
  volume    = {60},
  number    = {3},
  pages     = {448--476},
  doi       = {10.1093/ejcts/ezab038},
  pmid      = {34293102}
}

\end{document}